 \documentclass[twoside,reqno,11pt]{fcaa}

\usepackage{graphicx}
\usepackage{epsfig}
\usepackage{amsthm}
\usepackage{amsmath}
\usepackage{latexsym}
\usepackage{amsfonts}
\usepackage{amssymb}

\newtheoremstyle{theorem}
  {15pt}          
  {15pt}  
  {\sl}  
  {\parindent}
  {\sc}  
  {. }   
  { }    
  {}     
\theoremstyle{theorem}

\newtheoremstyle{defi}
  {15pt}          
  {15pt}  
  {\rm}  
  {\parindent}     
  {\sc}  
  {. }    
  { }    
  {}     
\theoremstyle{defi}

 \def\proofname{\indent\rm P\ r\ o\ o\ f.}
 
 \def\qed{\hfill$\Box$} 

\usepackage{amsmath}
\usepackage{amssymb}        
\usepackage{graphicx}
\numberwithin{equation}{section}
\numberwithin{figure}{section}
\usepackage{amsthm}
\usepackage{lscape}
\usepackage{arydshln}
\usepackage{slashbox}
\usepackage{subcaption}
\renewcommand{\figurename}{Fig.}
 \usepackage{hyperref} 

 \def\theequation{\arabic{section}.\arabic{equation}}

 \def\themycopyrightfootnote{\vspace*{3pt}
 \copyright \, Year\,  Diogenes  Co., Sofia
   \par  \noindent pp. xxx--xxx
   \hfill  \vspace*{-36pt}
  }

 \title[Fractional calculus in image processing: a review]{
 {\tt On the Occasion of Professor Richard L. Magin's 70th Birthday}  \vspace*{1cm} \\
 Fractional calculus in image processing: a review \\  }
 \author[Qi Yang, Dali Chen, Tiebiao Zhao, YangQuan Chen]{Qi Yang $^1$, Dali Chen $^2$, Tiebiao Zhao $^3$, YangQuan Chen $^4$}

\begin{document}


 \vbox to 2.5cm { \vfill }
\bibliographystyle{IEEEtran}

 \bigskip \medskip

 \begin{abstract}

Over the last decade, it has been demonstrated that many systems in science and engineering can be modeled more accurately by fractional-order than integer-order derivatives, and many methods are developed to solve the problem of fractional systems. Due to the extra free parameter order $\alpha $, fractional-order based methods provide additional degree of freedom in optimization performance. Not surprisingly, many fractional-order based methods have been used in image processing field. Herein recent studies are reviewed in ten sub-fields, which include image enhancement, image denoising, image edge detection, image segmentation, image registration, image recognition, image fusion, image encryption, image compression and image restoration. In sum, it is well proved that as a fundamental mathematic tool, fractional-order derivative shows great success in image processing.

 \medskip


 \smallskip


 \end{abstract}

 \maketitle

 \vspace*{-16pt}



 \section{Introduction}\label{sec:1}

\setcounter{section}{1}
\setcounter{equation}{0}\setcounter{theorem}{0}

The idea of fractional order derivative was mentioned in 1695 during discussions between Leibniz and L'Hospital: \textquotedblleft Can the meaning of derivatives with integer order be generalized to derivatives with non-integer orders?\textquotedblright. The question raised by Leibniz was ongoing in the past 300 years \cite{petravs2011fractional} till people like Liouville, Riemann, and Weyl made major contributions to the theory of fractional calculus.  Nowadays, people like Chen \cite{chen2002discretization}, Podlubny \cite{podlubny2000matrix}, Oustaloup \cite{oustaloup2000frequency}, Richard \cite{magin2006fractional,magin2008fractional,magin2008modeling,magin2008anomalous} and Xue \cite{dingyu2006control} extended the theory to mechanics, physics, control theory field, bioengineering and so on. Moreover, a few very good and interesting Matlab functions were programmed and already submitted to the MatlabCentral File Exchange, where they are available for sharing among the users.

The subject of fractional calculus and its applications have gained considerable popularity during the past decades or so in diverse fields of science and engineering, like dynamics system and image processing. As to image processing, the work flow using fractional-order is shown in the Fig.1.1. The flow contains three steps. Firstly, the effective operator, model or equation involving ordinary differentiation and integration is selected. Then, the ordinary differentiation and integration are generalized to fractional-order (arbitrary order) using the fractional calculus definition (G-L, R-L or Caputo). Finally, numerical approximation to the fractional-order operator, model or equation will be calculated by discretization methods.

\begin{figure}[h]
\centering
\includegraphics[width=\textwidth]{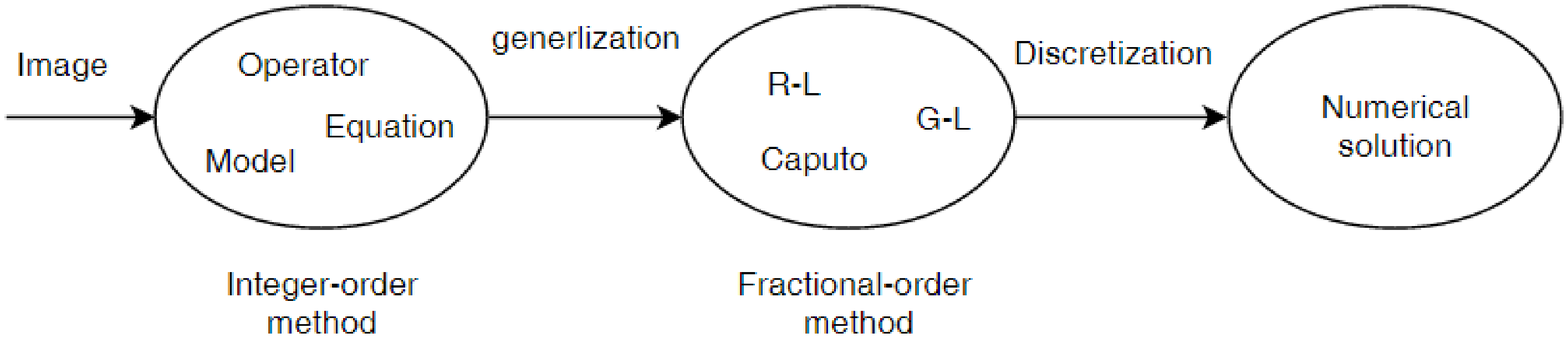}
\bigskip

Fig. 1.1: Fractional-order image processing flow
\label{fig:fractional-order image processing flow}
\end{figure}
The remainder of this paper is organized as follows: in Section \ref{sec:introduce}, fractional-order derivative definition, discretization method and related toolboxes in Matlab are introduced. Section \ref{sec:review} summarizes and analyzes fractional-order based image processing method in ten sub-fields, and demonstrates some experimental data. The paper is concluded in Section \ref{sec:conclusion}.

\section{Basics Fact of Fractional-order Derivative}
\label{sec:introduce}

\setcounter{section}{2}
\setcounter{equation}{0}\setcounter{theorem}{0}

According to the above flow, once we design the method involving differentiation or integration, the next step is to generalize it with fractional-order concept and finally obtain the numerical result. The fractional-order definition, approximating methods and toolboxes are introduced below.

 \subsection{Fractional-order Definition}\label{subsec:1.1}

Three popular definitions of fractional calculus were given by Gr\"{u}nwald-Letnikov (G-L), Riemann-Liouville (R-L), and Caputo. Of these, G-L and R-L are the most popular definitions used in digital image processing.\\
The G-L definition is defined as:
\begin{equation}\label{eq1}
_{a}^{ }\textrm{D}_{t}^{\alpha }=\lim_{h\to0}h^{-\alpha }\sum_{j=0}^{[\frac{t-a}{h}]}(-1)^{j}\binom{\alpha }{j}f(t-jh)
\end{equation}
where the $\binom{\alpha }{j}$ are the binomial coefficients. \\
The R-L definition is :
\begin{equation}\label{eq2}
_{a}^{ }\textrm{D}_{t}^{\alpha }=\frac{1}{\Gamma (n-\alpha )}\left ( \frac{d}{dt}\right )^{n}\int_{a}^{t}\frac{f(\tau )}{(t-\tau )^{\alpha -n+1}}d\tau\; \; \; \; \; \; \; (n-1<\alpha <n)
\end{equation}
where $\Gamma (.)$ is the gamma function, which is defined as:
\begin{equation}\label{eq4}
\Gamma (n)=\int_{0}^{\infty }t^{n-1}e^{-t}dt
\end{equation}
This function is a generalization of the factorial in the following form:
\begin{equation}\label{eq5}
\Gamma (n)=(n-1)!
\end{equation}
This gamma function is directly implemented with the function $gamma()$ in Matlab.\\
The Caputo definition of fractional derivatives is:
\begin{equation}\label{eq3}
_{a}^{ }\textrm{D}_{t}^{\alpha }=\frac{1}{\Gamma (n-\alpha )}\int_{a}^{t}\frac{f^{(n)}(\tau )}{(t-\tau )^{\alpha -n+1}}d\tau\; \; \; \; \; \; \; \; (n-1<\alpha <n)
\end{equation}
In fact, the three best known definitions G-L, R-L, and Caputo are equivalent under zero initial conditions.

\subsection{Fractional-order Discretization Methods and Toolboxes}
For numerical calculation of fractional-order derivatives we can use the formula derived from the G-L definition and matrix method.

\subsubsection{Gr\"{u}nwald-Letnikov Discretization Method}

Gr\"{u}nwald-Letnikov discretization method is given by
\begin{equation}\label{eq7}
_{a}^{ }\textrm{D}_{t}^{\alpha }f(t)=\lim_{h \to 0}\frac{1}{h^{\alpha }}\sum_{j=0}^{[\frac{t-a}{h}]}(-1)^{j}\binom{\alpha }{j}f(t-jh)\approx \frac{1}{h^{\alpha }}\sum_{j=0}^{[\frac{t-a}{h}]}\omega _{j}^{(\alpha )}f(t-jh)
\end{equation}
where $\omega_{j}^{(\alpha )}=(-1)^{j}\binom{\alpha }{j}$ is the polynomial coefficients of $(1-z)^{\alpha }$, and can be calculate by the following the recurrence formula:
\begin{equation}\label{eq6}
\omega _{0}^{(\alpha )}=1,\; \omega _{j}^{(\alpha )}=\left ( 1-\frac{\alpha +1}{j} \right )\omega _{j-1}^{(\alpha )},\; j=1,2,...
\end{equation}
The Matlab function $glfdiff()$ for the numerical approximate solution is given in \cite{dingyu2006control}.
\\
Another function $fgl\underline\;deriv(a,y,h)$ is provided in MatlabCentral \cite{Jonathan2014Fractional}.
\subsubsection{Matrix Discretization Method}
Suppose $t_{k} = kh (k = 0, 1, . . . ,N)$ are equidistant nodes with the step h in the interval [a, b], $t_{0}= a$ and $t_{N} = b$, according to the definition of formula (R-L) and using the backward fractional difference approximation for the $\alpha$-th derivative at the points $t_{k}$, k = 0, 1, . . . ,N, we have:
\begin{equation}\label{eq8}
_{a}^{ }\textrm{D}_{t_{k}}^{\alpha }f(t)\approx \frac{\triangledown ^{\alpha }f(t_{k})}{h^{\alpha }}=h^{-\alpha }\sum_{j=0}^{k}(-1)^{j}\binom{\alpha }{j}f_{k-j},\; \; \; \; k=1,2,...,N.
\end{equation}
All \emph{N} + 1 above formulas can be written simultaneously in the matrix form \cite{podlubny2009matrixartical}:

\begin{equation}\label{eq9}
\begin{bmatrix}
h^{-\alpha }\triangledown ^{\alpha }f(t_{0})\\
h^{-\alpha }\triangledown ^{\alpha }f(t_{1})\\
h^{-\alpha }\triangledown ^{\alpha }f(t_{2})\\
\vdots \\
h^{-\alpha }\triangledown ^{\alpha }f(t_{N-1})\\
h^{-\alpha }\triangledown ^{\alpha }f(t_{N})
\end{bmatrix}=\frac{1}{h^{\alpha }}\begin{bmatrix}
\omega _{0}^{(\alpha) } &  0&  0&  0&  \cdots& 0\\
 \omega _{1}^{(\alpha) }&  \omega _{0}^{(\alpha) }&  0&  0&  \cdots& 0\\
 \omega _{2}^{(\alpha) }&  \omega _{1}^{(\alpha) }&  \omega _{0}^{(\alpha) }&  0&  \cdots& 0\\
 \ddots &   \ddots &   \ddots &   \ddots &  \cdots & \cdots\\
 \omega _{N-1}^{(\alpha) }&  \ddots &  \omega _{2}^{(\alpha) }&  \omega _{1}^{(\alpha) }&  \omega _{0}^{(\alpha) }& 0\\
 \omega _{N}^{(\alpha) }&  \omega _{N-1}^{(\alpha) }& \ddots  &  \omega _{2}^{(\alpha) }&  \omega _{1}^{(\alpha) }& \omega _{0}^{(\alpha) }
\end{bmatrix}\begin{bmatrix}
f_{0}\\
f_{1}\\
f_{2}\\
\vdots \\
f_{N-1}\\
f_{N}
\end{bmatrix}
\end{equation}
\begin{equation}\label{eq10}
\omega _{j}^{(\alpha )}=(-1)^{j}\binom{\alpha }{j},\; \; \; j=1,2,...N.
\end{equation}
\begin{equation}\label{eq11}
B_{N}^{\alpha }=\frac{1}{h^{\alpha }}\begin{bmatrix}
\omega _{0}^{(\alpha) } &  0&  0&  0&  \cdots& 0\\
 \omega _{1}^{(\alpha) }&  \omega _{0}^{(\alpha) }&  0&  0&  \cdots& 0\\
 \omega _{2}^{(\alpha) }&  \omega _{1}^{(\alpha) }&  \omega _{0}^{(\alpha) }&  0&  \cdots& 0\\
 \ddots &   \ddots &   \ddots &   \ddots &  \cdots & \cdots\\
 \omega _{N-1}^{(\alpha) }&  \ddots &  \omega _{2}^{(\alpha) }&  \omega _{1}^{(\alpha) }&  \omega _{0}^{(\alpha) }& 0\\
 \omega _{N}^{(\alpha) }&  \omega _{N-1}^{(\alpha) }& \ddots  &  \omega _{2}^{(\alpha) }&  \omega _{1}^{(\alpha) }& \omega _{0}^{(\alpha) }
\end{bmatrix}
\end{equation}
We can consider the matrix $B_{N}^{\alpha }$ as a discrete analogue of left-sided fractional differentiation of  $\alpha $-order, and the right-sided fractional derivative is defined as:
\begin{equation}\label{eq11-4}
_{t}^{ }\textrm{D}_{b}^{\alpha }=\frac{(-1)^{n}}{\Gamma (n-\alpha )}\left ( \frac{d}{dt}\right )^{n}\int_{t}^{b}\frac{f(\tau )}{(\tau-t )^{\alpha -n+1}}d\tau\; \; \; \; \; \; \; (n-1\leq \alpha <n,a<t<b)
\end{equation}
Similarly to the left-sided fractional differentiation, numerical solution of the right-sided fractional differentiation is represented by the matrix \cite{podlubny2009matrixartical}:
\begin{equation}\label{eq11-5}
F_{N}^{\alpha }=\frac{1}{h^{\alpha }}\begin{bmatrix}
\omega _{0}^{(\alpha) } &  \omega _{1}^{(\alpha) }&  \ddots&  \ddots& \omega _{N-1}^{(\alpha) }&   \omega _{N}^{(\alpha) }\\
 0&  \omega _{0}^{(\alpha) }&  \omega _{1}^{(\alpha) }&  \ddots&  \ddots& \omega _{N-1}^{(\alpha) }\\
 0&  0&  \omega _{0}^{(\alpha) }&  \omega _{1}^{(\alpha) }&  \ddots& \ddots\\
 \cdots&   \cdots &   \cdots &   \ddots &  \ddots & \ddots\\
 0&  \cdots &  0&  0&  \omega _{0}^{(\alpha) }& \omega _{1}^{(\alpha) }\\
 0&  0& \cdots  &  0&  0& \omega _{0}^{(\alpha) }
\end{bmatrix}
\end{equation}
According to left-sided fractional differentiation, the corresponds to right-sided differentiation as:
\begin{equation}\label{eq11-6}
(B_{N}^{\alpha })^{T}=F_{N}^{\alpha }
\end{equation}
The function $fracdiffdemoy(alpha,beta)$ is provided for matrix numerical solution in the MatlabCentral \cite{Podlubny2009Matrix}.
\begin{figure}[h]
\centering
\includegraphics[width=0.7\textwidth]{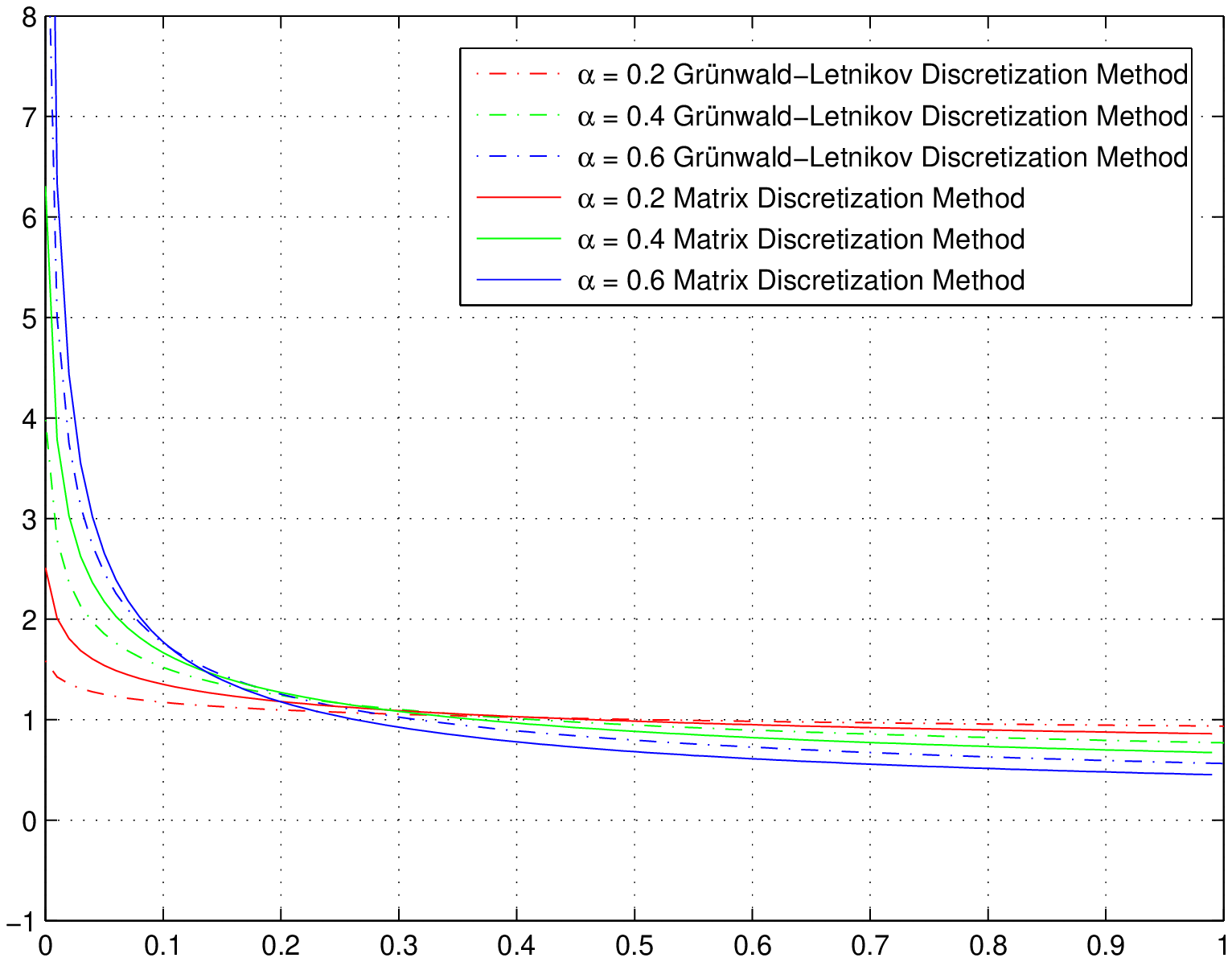}
\bigskip

Fig. 2.1: Comparison of Grünwald-Letnikov discretization method and matrix discretization method \cite{podlubny2009matrixartical}
 \label{fig:Matrix}
\end{figure}
\\
Fig.2.1 compare right-sided Riemann-Liouville fractional derivatives and Gr\"{u}nwald-Letnikov fractional derivatives of the function $y(x)=1$ with the orders $\alpha$ as 0.2, 0.4 and 0.6.

\section{A Review of Fractional-order Image Processing Methods}
\label{sec:review}
This section mainly introduces and analyzes the fractional-order based image processing methods, meanwhile summarizes and classifies the existing literature involving fractional-order image processing, and gives part of the simulation results.
\subsection{Image Enhancement}
As for image enhancement, most integral differential operators work well when used for high-frequency features of image (e.g. Sobel, Prewitt, and Laplacian of Gaussian operators). Nevertheless, their performance deteriorates significantly when applied to smooth regions. Whereas the fractional differential operator has the capability of not only preserving high-frequency contour features, but also improving the low-frequency texture details in smooth area. Therefore, more and more fractional differentiation-based methods were applied in the field of image enhancement. A number of cutting-edge techniques have been proposed in two categories: transform-based \cite{li2013image, roy2015fractional} and spatial domain-based \cite{li2016fractional, pu2010fractional, chen20121, li2015adaptive, li2015image, hu2015adaptive, si2014texture, che2012fractional}. In transform-based methods, images are converted to fractional frequency domain, and the coefficients of filter function are regulated. Finally, all the output images are obtained by inverse transform. These methods improve the image contrast to achieve better texture and seldom noise is introduced, whereas spatial domain-based methods using fractional differential approaches can completely avoid noise introduction while enhancing image. Among the different fractional differential approaches, fractional differential mask operator design stands out as a particularly important method.
\begin{figure}[h]
\centering
\includegraphics[width=0.75\textwidth]{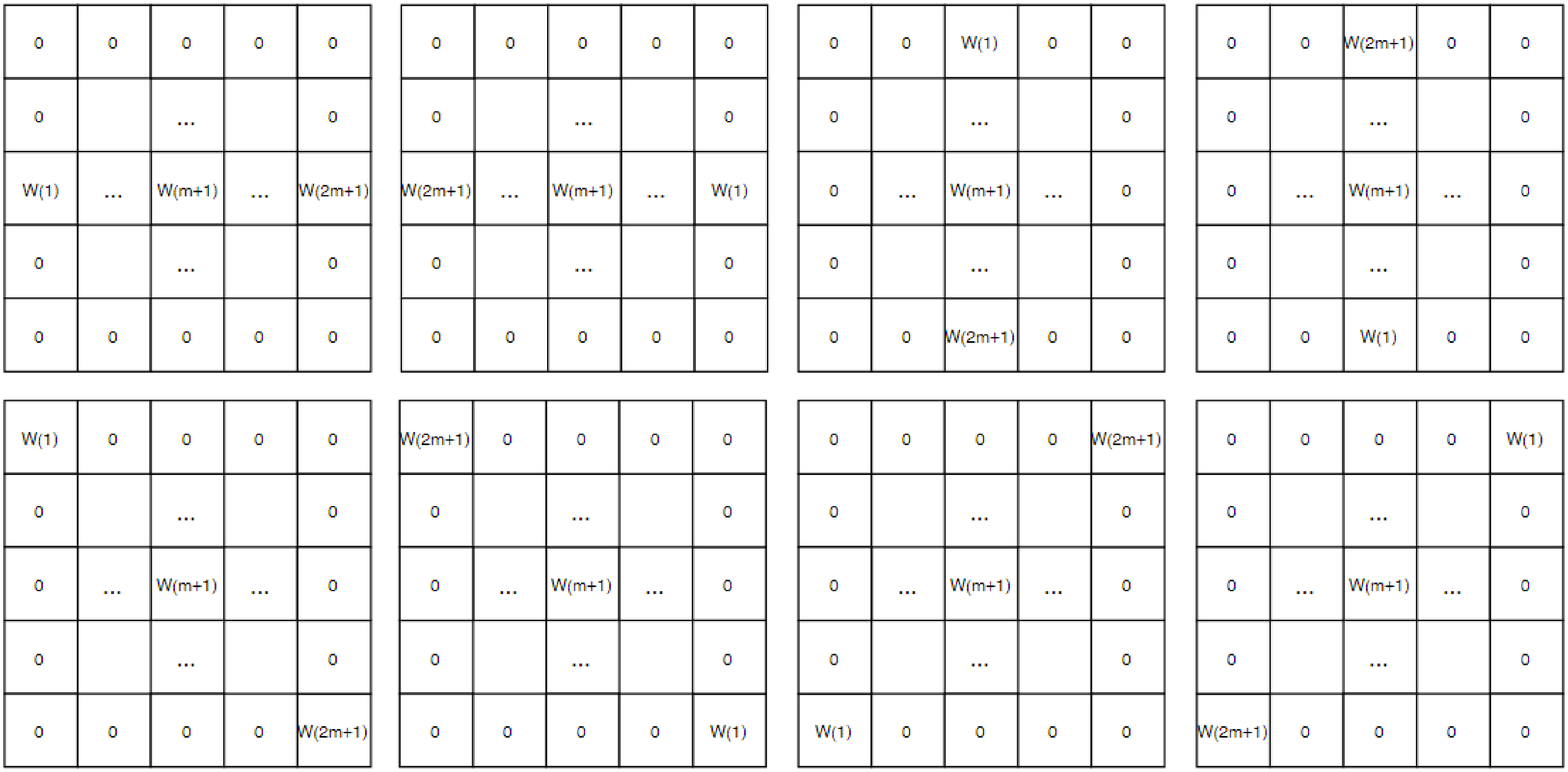}
\bigskip

Fig. 3.1: Eight directions operator \cite{chen20121}
\label{fig:Eight directions operator}
\end{figure}\\

Commonly fractional differential mask operators are designed or rewritten for the purposes of fractional differential, fractional integral, different parameter range, best convergence and precision respectively \cite{li2016fractional, pu2010fractional}. Mask structure includes single direction, multi-direction (group direction) and non-regular region. Single direction masks are difficult to capture image gradient for better enhancement result. To obtain the fractional differential on the multiple symmetric directions and make the fractional differential masks have anti-rotation capability, 8 fractional differential masks are designed which include the directions of negative x-coordinate, negative y-coordinate, positive x-coordinate, positive y-coordinate, left downward diagonal, right upward diagonal, left upward diagonal, and right downward diagonal \cite{pu2010fractional, chen20121, li2015adaptive}(see Fig.3.1). The $\alpha $th-order derivatives of \emph{G}(\emph{x}, \emph{y}) in the different directions are calculated by equation \eqref{eq12} , equation \eqref{eq13} and an unsupervised optimization algorithm is used for choosing the fractional order in \cite{chen20121} (see Fig.3.2, Fig.3.3). To improve the anti-rotation performance of algorithm, the 16 directions  of the gradient operator is used to construct a $5\times 5$ fractional differential mask \cite{li2015image}. For non-regular region enhancement, fractional differential operator with non-integer step and order is used to enhance image texture \cite{hu2015adaptive}. By constructing a non-regular self-similar support region according to a local texture similarity measure it can effectively exclude pixels with low correlation and noise. In the image, edge and texture characteristics are different with \emph{a}th-order value. To treat the edge and texture differently, piecewise function of $\alpha $th-order fractional differential was used to design the corresponding adaptive fractional differential function, with a high order in edge pixels and a relatively small order in the weak texture pixels \cite{si2014texture, che2012fractional}. It can make edges clearer and preserve weak textures. The 8 directions fractional differential masks are defined as Fig.3.1.
\begin{figure}[htbp]
  \centering
   \begin{subfigure}[b]{0.25\textwidth}
   \centering
        \includegraphics[width=1.2in]{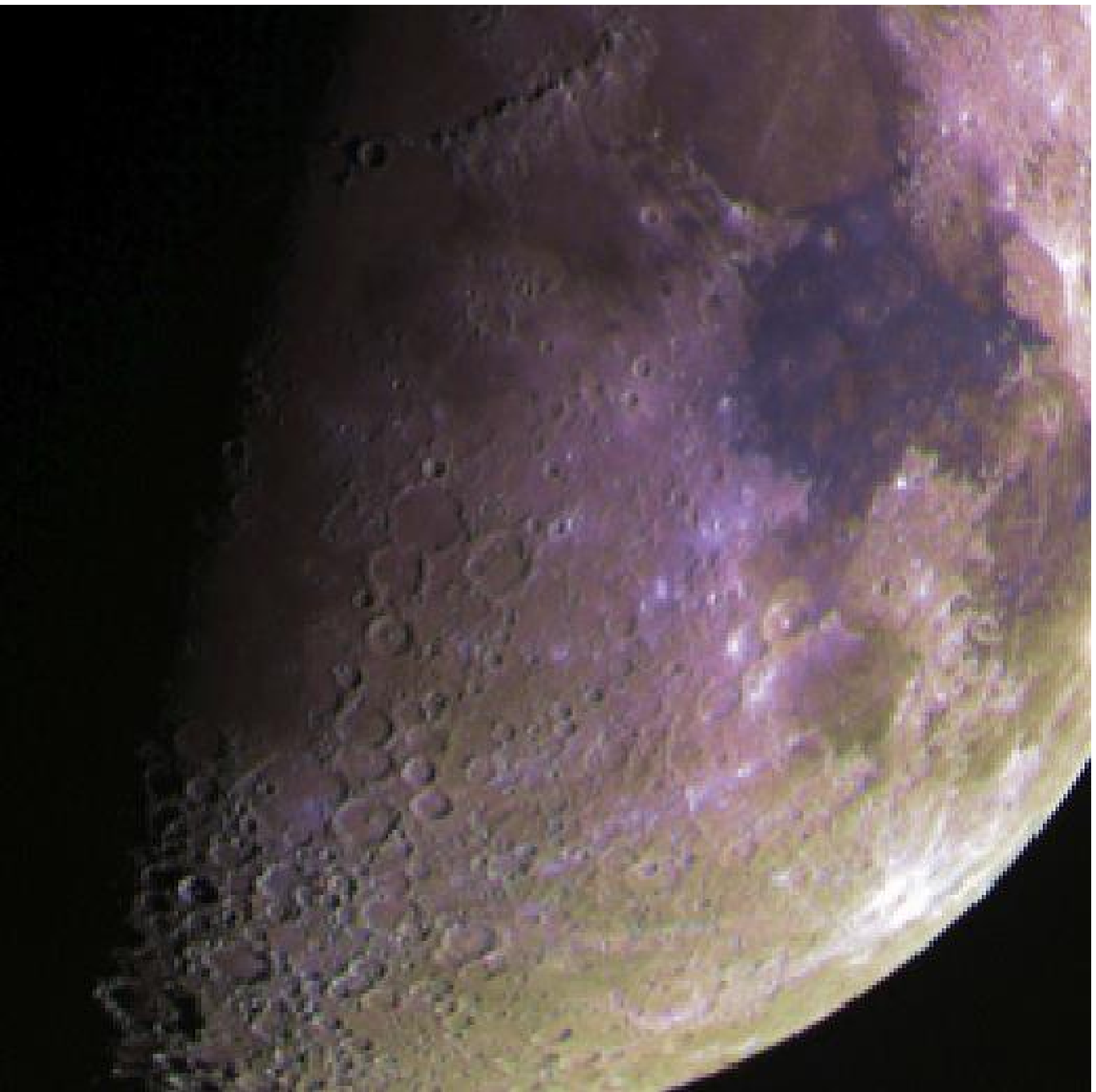}
        \caption{{\tt Original}}
    \end{subfigure}%
   \begin{subfigure}[b]{0.25\textwidth}
   \centering
        \includegraphics[width=1.2in]{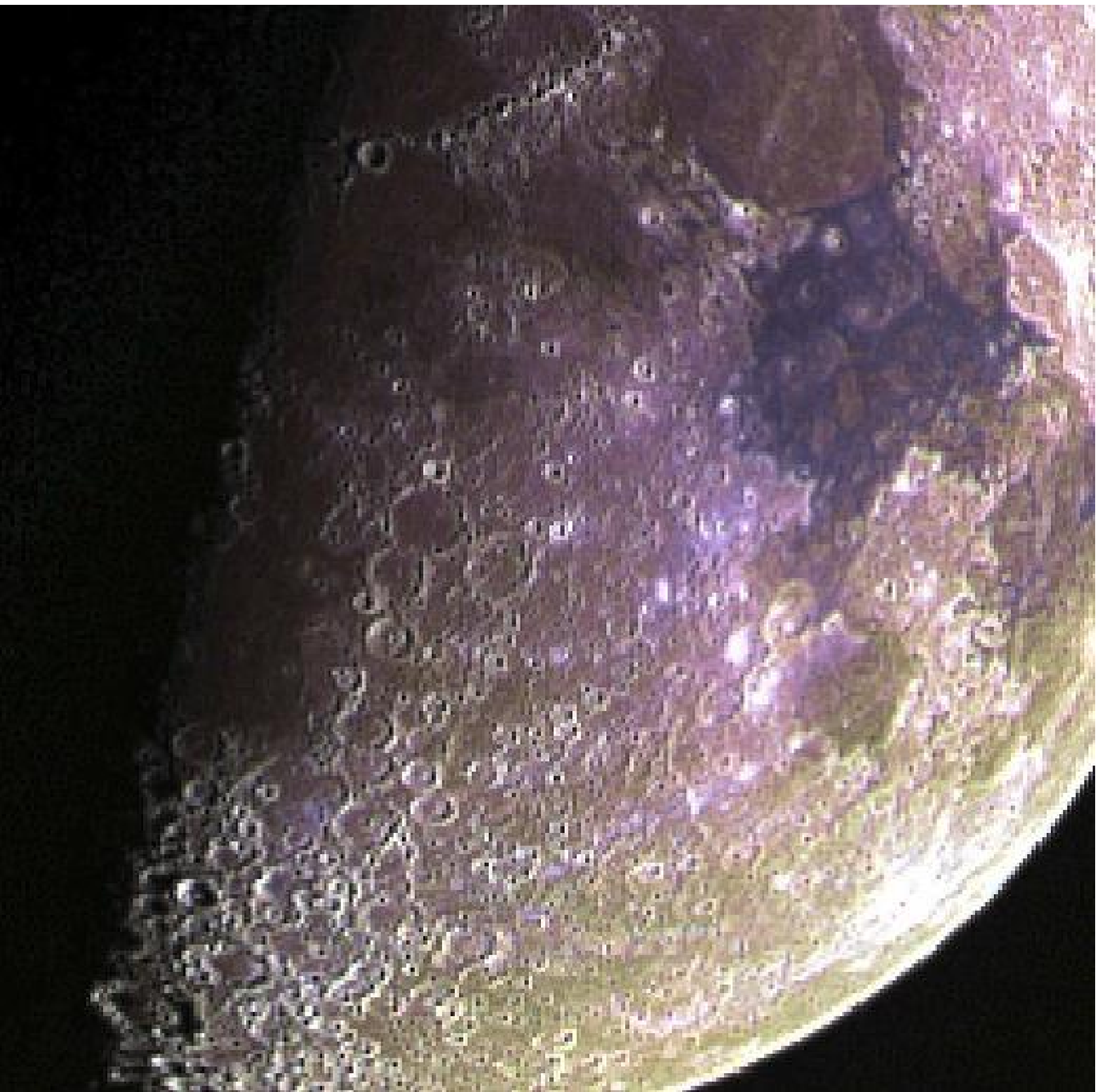}
        \caption{{\tt  $\alpha $ = 0.53}}
   \end{subfigure}%
     \centering
    \begin{subfigure}[b]{0.25\textwidth}
    \centering
         \includegraphics[width=1.2in]{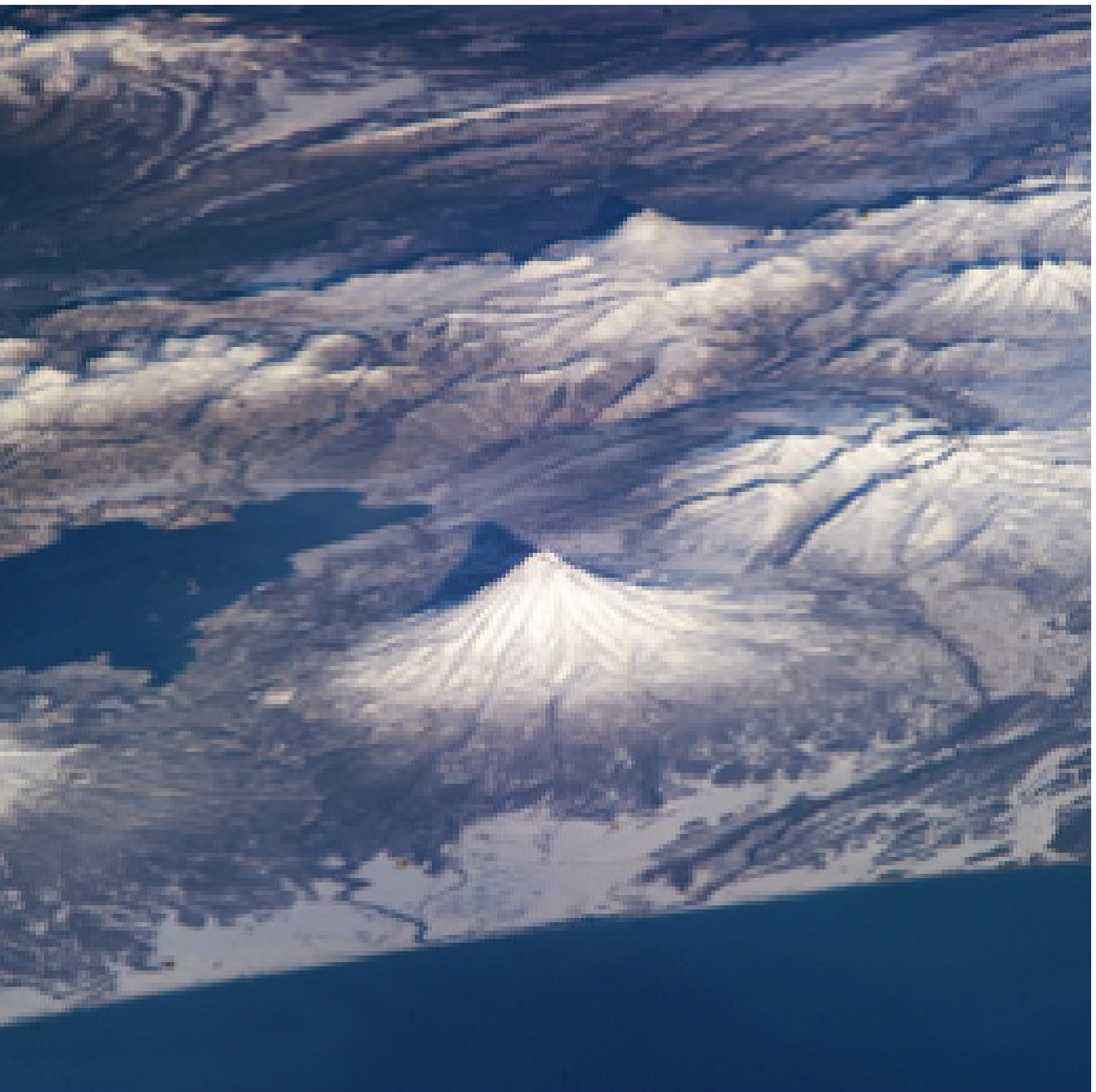}
         \caption{{\tt  Original }}
     \end{subfigure}%
    \begin{subfigure}[b]{0.25\textwidth}
    \centering
         \includegraphics[width=1.2in]{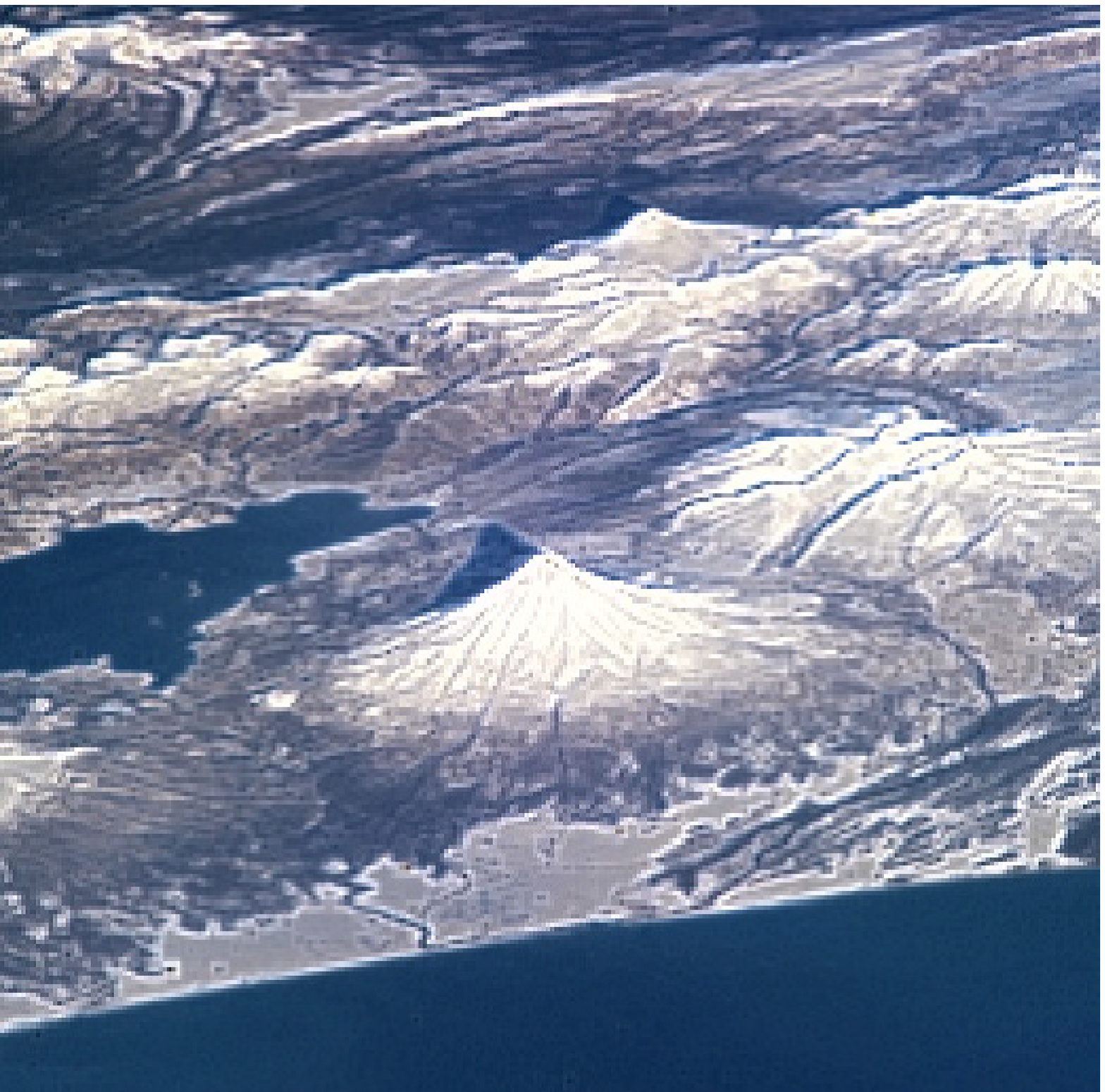}
         \caption{{\tt  $\alpha $ = 0.47}}
    \end{subfigure}%
    \\
    \bigskip

   Fig. 3.2: Comparison between the original image and the enhanced image\cite{chen20121}
    \label{fig:Eight directions result}
\end{figure}
Fig.3.2 compares the original image with the enhanced image using eight directions mask, and we can see that the detail texture is enhanced greatly. Fig.3.3 adopts the adaptive fractional-order selection method for the best texture. The optimized $\alpha $ is 0.28 that are better than others enhanced images.
\begin{equation}\label{eq12}
W_{m+1}^{(\alpha)}=[W(1),W(2),...,W(2m+1))];
\end{equation}

\begin{eqnarray}\label{eq13}
\begin{split}
    G_{x^{+}}^{(\alpha)}(x,y)=\sum_{k=-m}^{m}\sum_{l=-m}^{m}W_{x^{+}}^{(\alpha)}(k,l)G(x-k,y-l);\\
    G_{x^{-}}^{(\alpha)}(x,y)=\sum_{k=-m}^{m}\sum_{l=-m}^{m}W_{x^{-}}^{(\alpha)}(k,l)G(x-k,y-l);\\
    G_{y^{+}}^{(\alpha)}(x,y)=\sum_{k=-m}^{m}\sum_{l=-m}^{m}W_{y^{+}}^{(\alpha)}(k,l)G(x-k,y-l);\\
    G_{y^{-}}^{(\alpha)}(x,y)=\sum_{k=-m}^{m}\sum_{l=-m}^{m}W_{y^{-}}^{(\alpha)}(k,l)G(x-k,y-l);\\
    G_{LDD}^{(\alpha)}(x,y)=\sum_{k=-m}^{m}\sum_{l=-m}^{m}W_{LDD}^{(\alpha)}(k,l)G(x-k,y-l);\\
    G_{RUD}^{(\alpha)}(x,y)=\sum_{k=-m}^{m}\sum_{l=-m}^{m}W_{RUD}^{(\alpha)}(k,l)G(x-k,y-l);\\
    G_{LUD}^{(\alpha)}(x,y)=\sum_{k=-m}^{m}\sum_{l=-m}^{m}W_{LUD}^{(\alpha)}(k,l)G(x-k,y-l);\\
    G_{RDD}^{(\alpha)}(x,y)=\sum_{k=-m}^{m}\sum_{l=-m}^{m}W_{RDD}^{(\alpha)}(k,l)G(x-k,y-l);\\
\end{split}
\end{eqnarray}

\begin{figure}[htbp]
   \begin{subfigure}[b]{0.25\textwidth}
   \centering
        \includegraphics[width=1.2in]{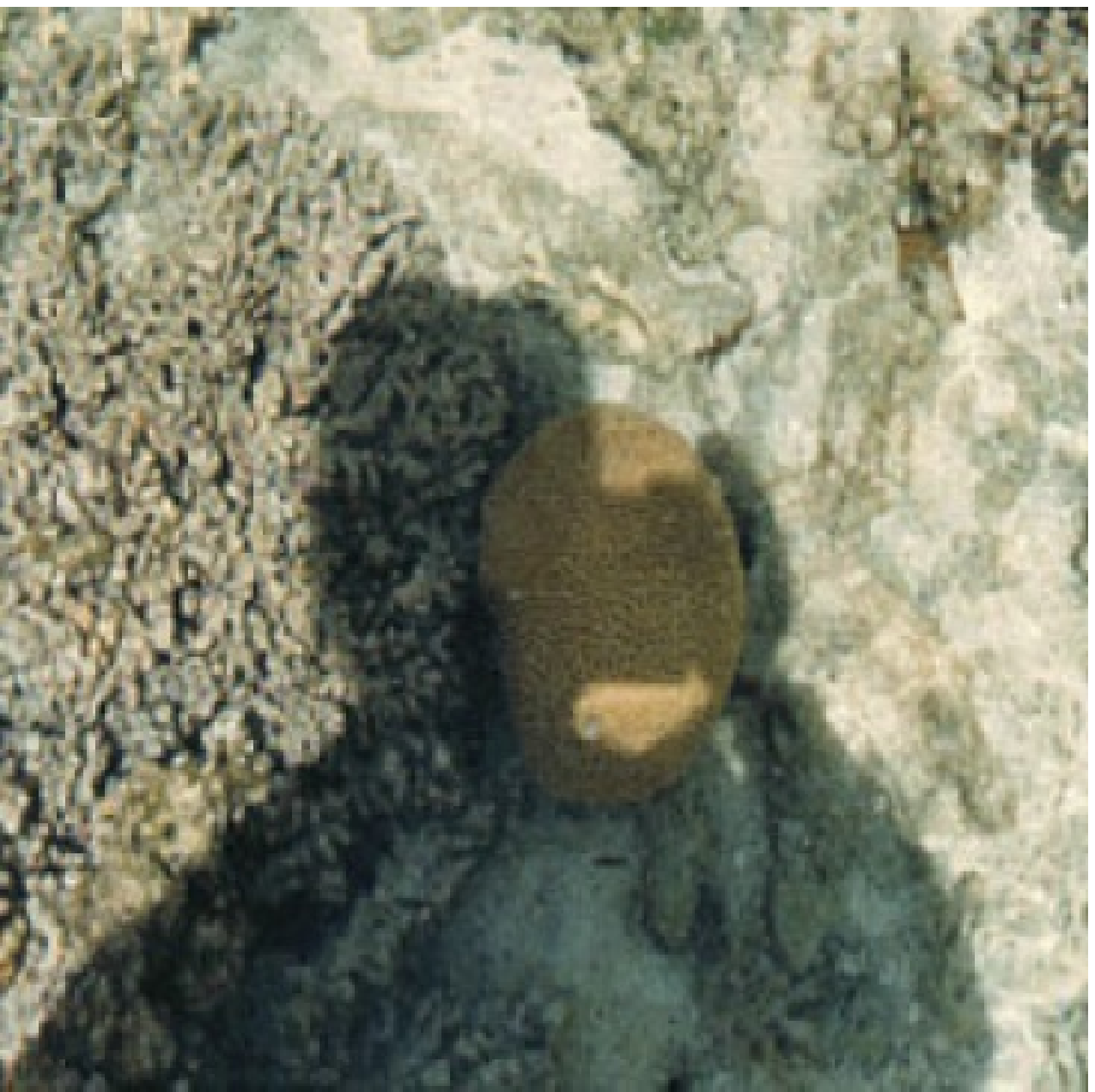}
        \caption{{\tt  $\alpha $ = 0.1}}
    \end{subfigure}%
   \begin{subfigure}[b]{0.25\textwidth}
        \includegraphics[width=1.2in]{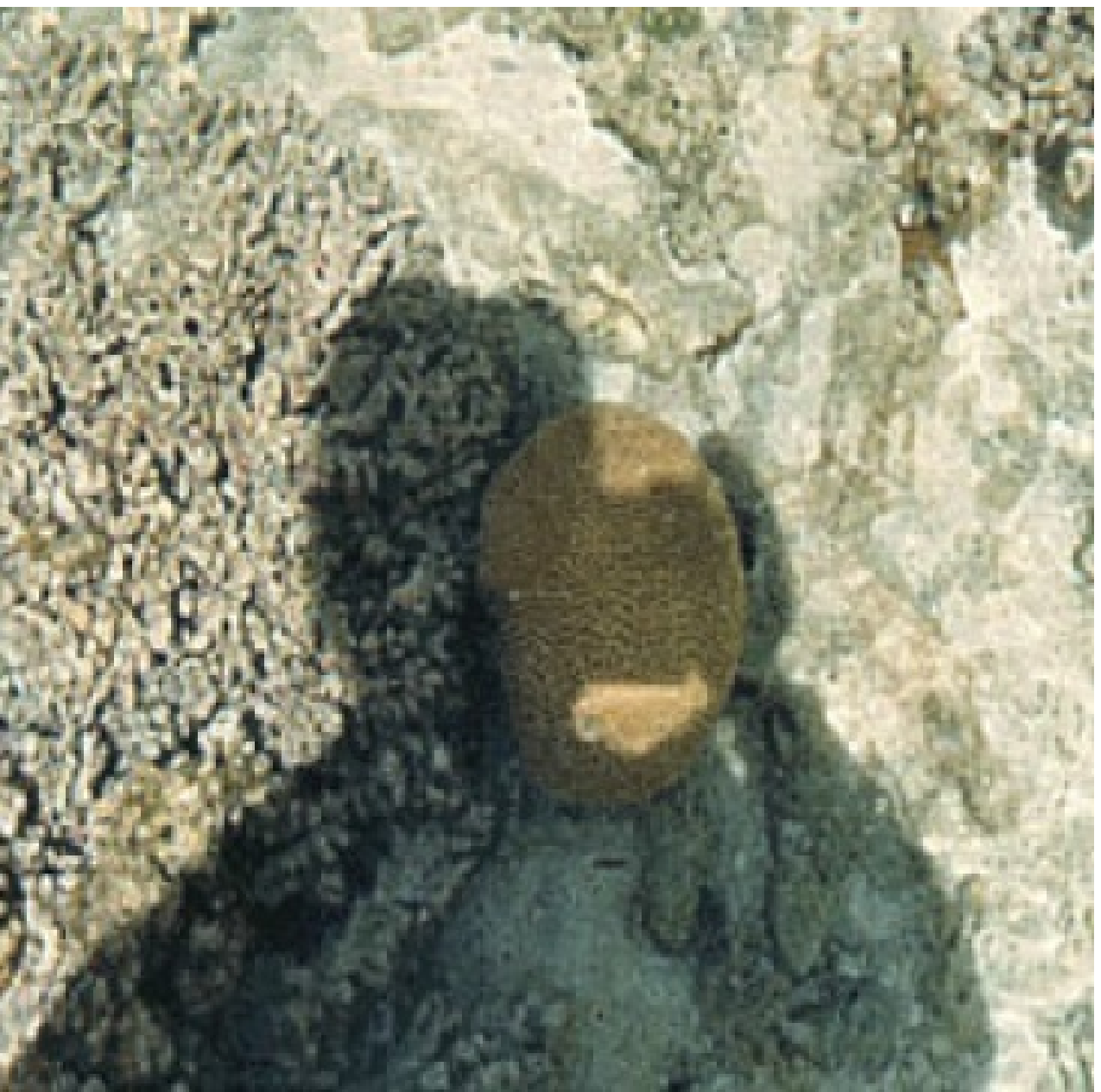}
        \caption{{\tt  $\alpha $ = 0.28}}
   \end{subfigure}%
   \begin{subfigure}[b]{0.25\textwidth}
        \includegraphics[width=1.2in]{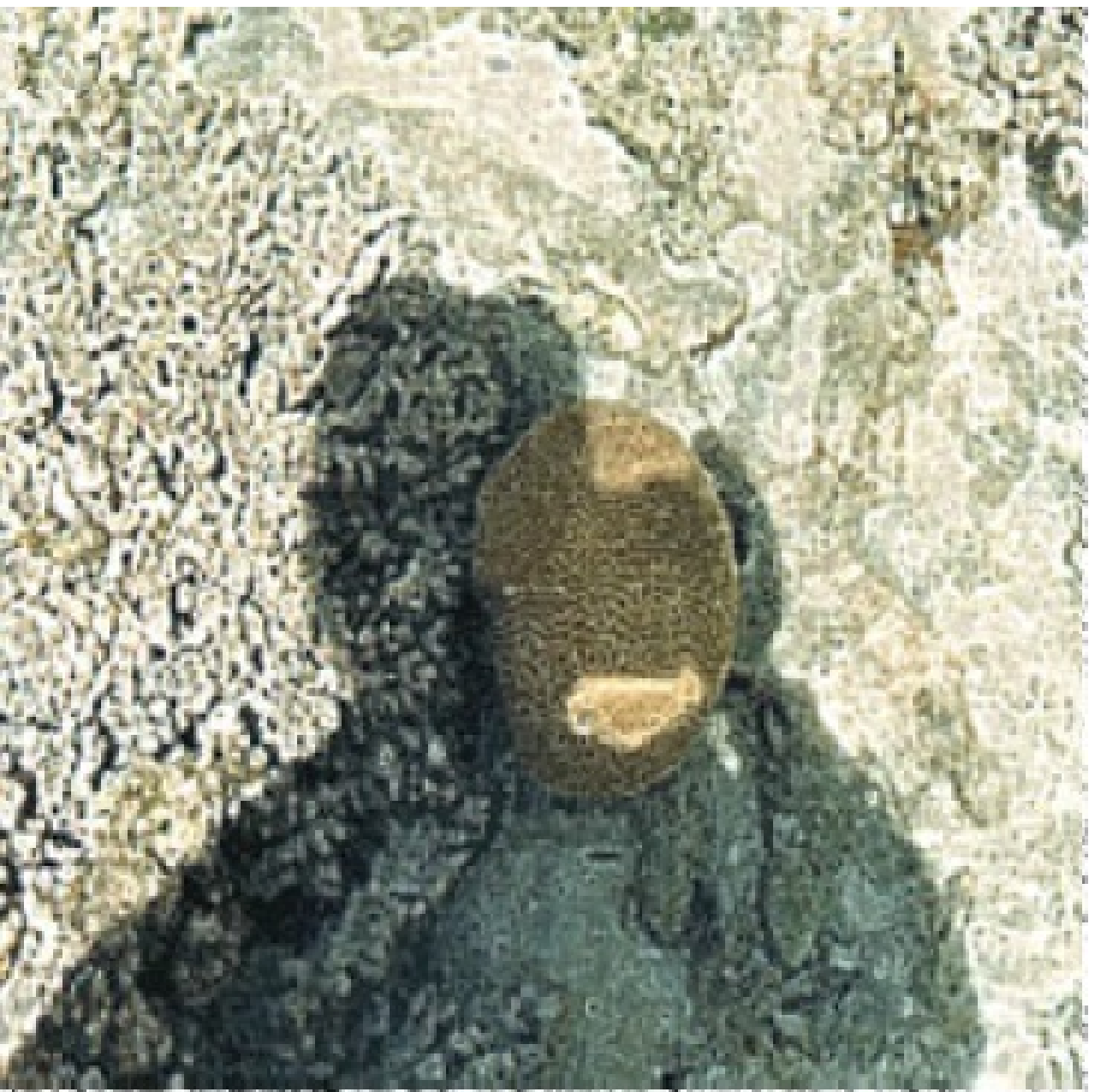}
        \caption{{\tt  $\alpha $ = 0.5}}
   \end{subfigure}%
   \begin{subfigure}[b]{0.25\textwidth}
        \includegraphics[width=1.2in]{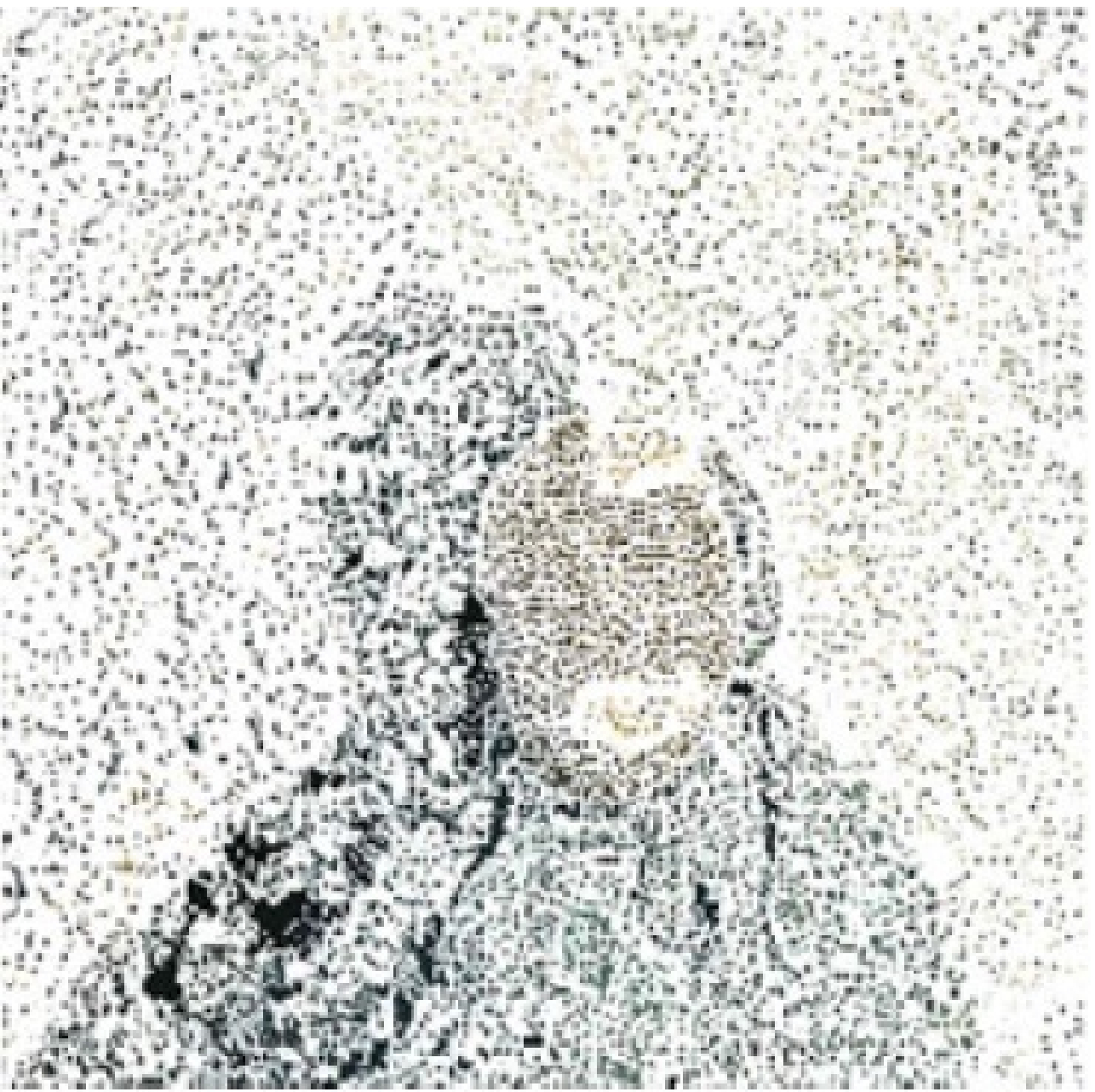}
        \caption{{\tt  $\alpha $ = 0.9}}
   \end{subfigure}%
   \bigskip

   Fig. 3.3: Comparison between four enhanced images with different $\alpha $'s \cite{chen20121}
  \label{fig:adapive_selection}
\end{figure}

\subsection{Image Denoising}

It is very easy to introduce noise in the process of capturing and transmitting images, hence it becomes necessary to reduce the noise level to preserve the image texture or edge features. Currently, fractional calculus provides an important tool for image denoising, which can be divided into three categories: operator mask-based \cite{li2016image,he2015improved,jalab2015fractional,che2013image}, model-based \cite{jun2011class,tian2015fractional,hu2012new,dong2013variational,chen2015fractional,chen2013fractional,chen2016robust,zhang2015total,brito2010multigrid,chan2010multilevel,chan2006optimization,zhang2012iterative,zachevsky2016statistics} and anisotropic diffusion equation based \cite{zhang2014fractional,bai2007fractional,zhang2015spatial}. Operator mask-based methods use fractional order differentiation or integration to construct fractional operator in advance (e.g. Alexander polynomials, Srivastava Owa and fractional partial differential), based on which the structures of $ n\times n $ fractional mask windows on symmetrical directions are constructed. The window size is usually set as $3\times 3$ or $5\times 5$ pixels in eight directions (see Fig.3.1).  The methods could remove noise while preserving texture and edge detail of image (see Fig.3.4). However, these methods focus on local and it is difficult to achieve the global optimum.
In order to obtain the global optimum, many fractional model-based methods have been proposed. These fractional model-based algorithm construct the observed image \emph{f} as a sum of two parts, e.g. $f = u + v$, where \emph{u} is a piecewise smooth part representing the clean image and \emph{v} is the oscillatory part representing the noise and textures. According to above theory, the minimization equation in integer-order derivative have been proposed by Rudin, Osher and Fatemi \cite{rudin1992nonlinear}.

\begin{equation}\label{eq14}
\underset{u\in BV}{min}\left \{ F(u)=\left | u \right |_{BV}+\frac{\lambda }{2}\left \| f-u \right \|_{2}^{2} \right \}
\end{equation}

where $\left | u \right |_{BV}$ represents the total variation of \emph{u} named as regularization term, $BV$ is the functions space, $\lambda $ is the regularization parameter and $\left \| f-u \right \|_{{2}}^{2}$ is a fidelity term. The equation \eqref{eq14} is known as the ROF model. It performs very well on noise reducing while maintaining edges and texture. However, the model has two defects:\\
(i) The ROF integer-order derivative model may lead to the staircase effect (or called blocky effect) since piecewise constant solution is adopted. \\
(ii) The ROF integer-order derivative model cannot maintain better textures. The noise and textures are modelized as the same oscillatory component and cannot be separated. An experiment shows that the model cannot separate  textures and noise with different oscillatory components of different frequencies \cite{gilles2010properties} , and often removes the textures along with noise in the process of denoising \cite{jun2011class}.

With respect to this problem, many methods using fractional-order derivative in the regularized term of the ROF model are proposed. It has been proved that the fractional-order derivative serves the biological visual system better than the integer-order derivative. \cite{jun2011class,pu2006fractional,yi2007fractional} show that the fractional-order derivative not only maintains the contour feature in the smooth area, but also preserves high-frequency components like edges and textures.
To eliminate the undesirable staircase effect, fractional anisotropic diffusion equation methods are proposed. As generalizations of second-order and fourth-order anisotropic diffusion equations \cite{bai2007fractional}, the Euler-Lagrange is redefined as an increasing function of the fractional derivative of the image intensity function. In order to improve its ability of wave propagation for image denoising, fractional diffusion-wave equation and spatial fractional telegraph equation are used to obtain an integrated behavior of diffusion and wave propagation, that is, to preserve edges or texture in a highly oscillatory region \cite{zhang2014fractional,zhang2015spatial}.  These equations have good visual effects and better signal-to-noise ratio.
Fig.3.4 shows image denoising in different parameter mask \emph{M}. From the picture we can show that the bigger \emph{M} results in better smoothing and denoising performance.
\begin{figure}[htbp]

   \begin{subfigure}[b]{0.3\textwidth}

        \includegraphics[width=1.3in]{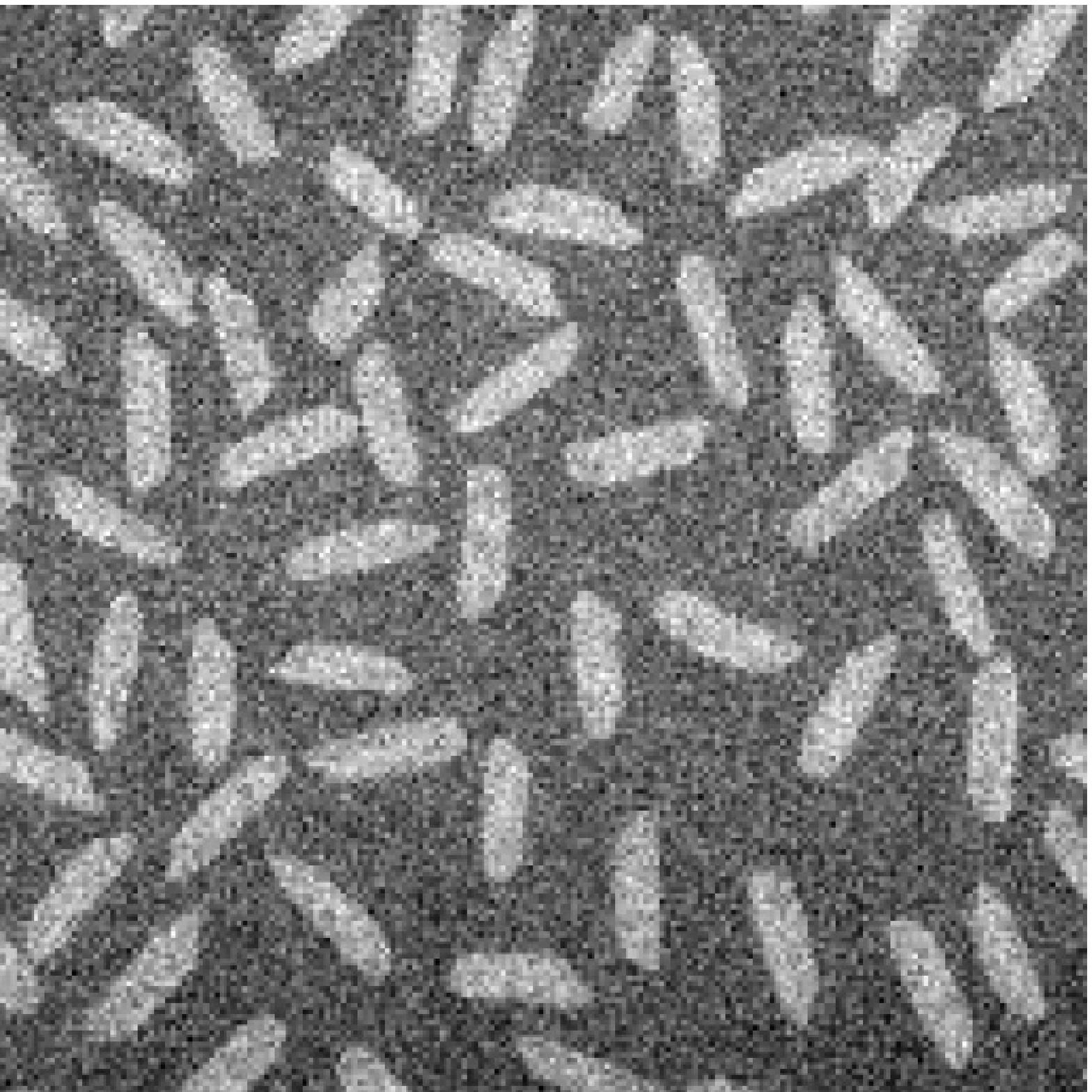}
        \caption{{\tt Original image}}
    \end{subfigure}%
   \begin{subfigure}[b]{0.7\textwidth}

        \includegraphics[width=3.4in]{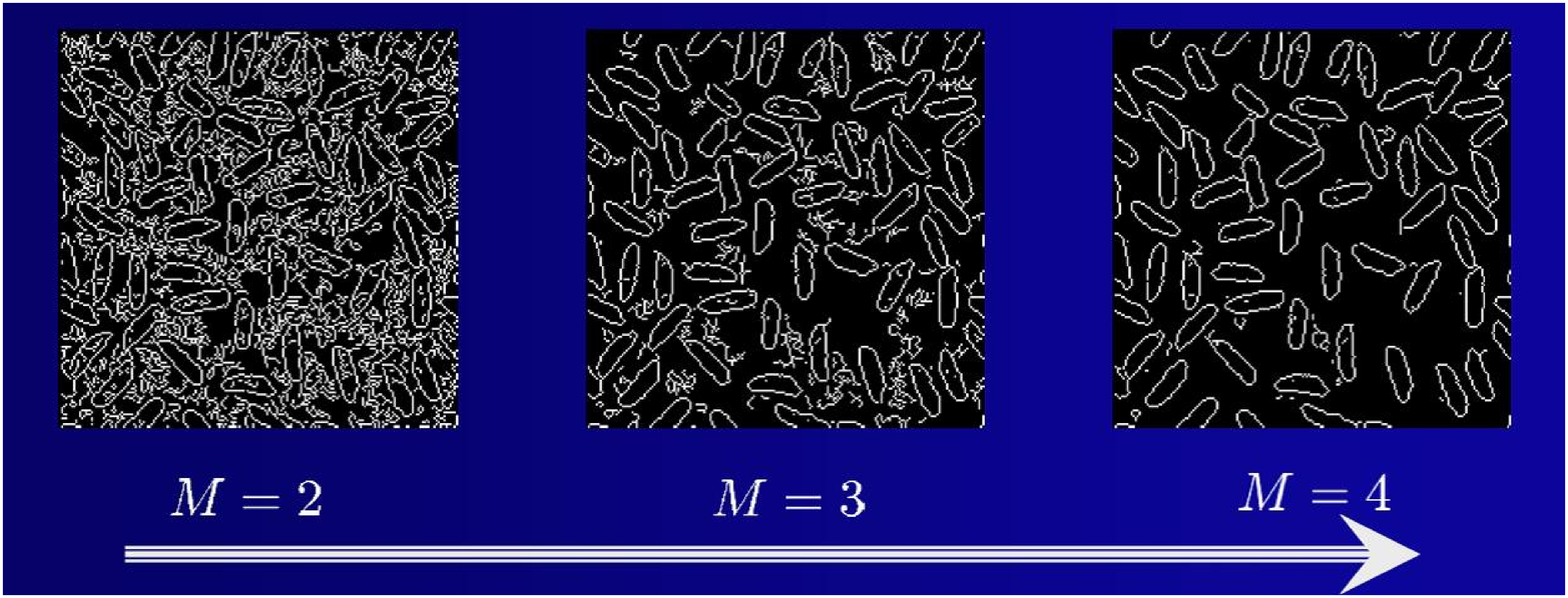}
        \caption{{\tt Fractional-order denoising}}
   \end{subfigure}%
    \bigskip

   Fig. 3.4: Fractional-order image denoising in different parameter mask \emph{M}
  \label{fig:image_denoising}
\end{figure}

\subsection{Image Edge Detection}
In image processing, integer-order differentiation operators are often used in edge detection, especially first order for the gradient (e.g. Roberts, Prewitt and Sobel) and second order for the Laplacian (e.g. Laplacian of Gaussian). However, the first order derivative methods generally cause thicker edges, resulting in the loss of image details. The second order derivative methods have a stronger response to fine detail, but they are more sensitive to noise. To solve this problem, the fractional order derivative has been introduced in the edge detection methods, with the capability to preserve more low-frequency contour features in the smooth areas, maintain high-frequency marginal features and also enhance medium-frequency texture details. Many fractional order operators are used for edge detection, such as fractional order Sobel operator \cite{tian2014fractional}, fractional order CRONE operator \cite{yang2010novel,mathieu2003fractional}, quaternion fractional differential operator \cite{gao2011edge}, fractional Laplace transform \cite{pan2013novel}, and Newton interpolation's fractional differentiation operator \cite{gao2014edge}. Chen \cite{chen2012Fractionalrobust} expands Riemann-Liouville fractional integral definition from one-dimensional to two-dimensional to analyze the performance of smoothing and enhancing by tuning the frequency-domain $\alpha $ and parameter mask width \emph{M}. The method simulation shows that the bigger $\alpha $ results in worse smoothing and better enhancing, the bigger \emph{M} works on the contrary to $\alpha $(see Fig.3.4). False reject rate (FRR), false accept rate (FAR) and single pixel detecting (SPD) are used to evaluate the method. Fig.3.5 compares fractional edge detection method with other traditional methods. It shows that the fractional method obtains sharper edge and better edge recognition rate than the other methods.

\begin{figure}[h]
\centering
\includegraphics[width=0.8\textwidth]{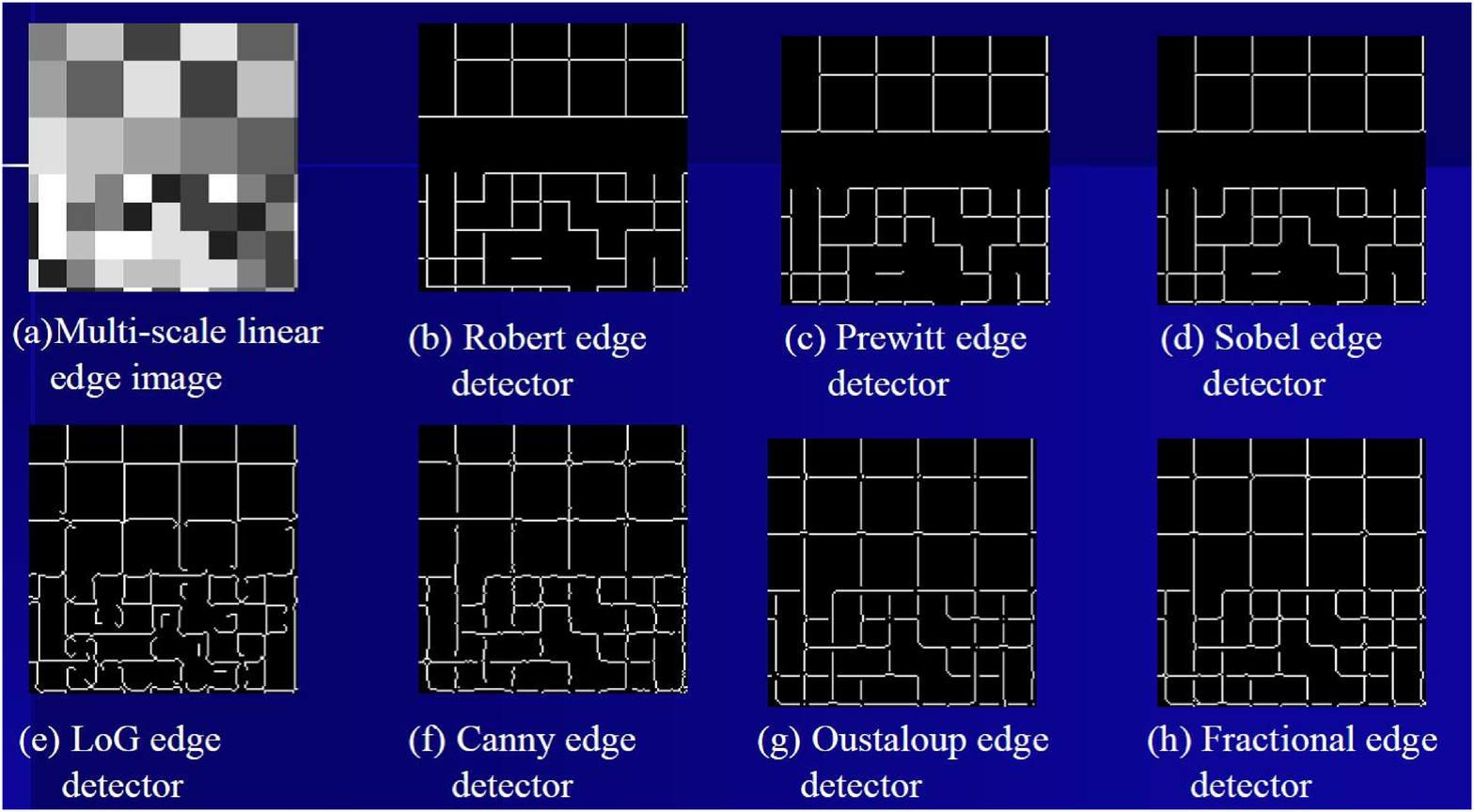}
\bigskip

   Fig. 3.5: Fractional-order method and other method comparison result
\label{fig:edge-2}
\end{figure}

\subsection{Image Segmentation}
The goal of image segmentation is to divide an image into its constituent component that is more meaningful and easier to analyze. Up to now, a broad variety of models have been proposed to solve the image segmentation problem, in which geometric active contour (AC) model using the level set method is very efficient. It segments objects images using dynamic curves based on an energy-minimizing model. However, it is sensitive to initial conditions and hard to deal with topological changes, like the combining and separating of the evolving curve. In order to solve this problem, the regularization is necessary to diminish the influence of noise and smooth the curve. So a new adaptive active contour model based on fractional order differentiation is proposed to realize it \cite{tian2013fractional,ren2015adaptive,zhang2014local}. This method generalizes the model regulation term with fractional derivative of the level set function instead of the integral order gradient. The fractional order gradient term shows better capability of preserving texture and extracting more image details \cite{tian2013fractional}. The other method combines the fractional order fitting term with the global fitting term to form a novel fitting term to describe the original image more accurately, and is robustness to noise \cite{ren2015adaptive}. Region scalable fitting(RSF) model is generalized with fractional-order gradient to get better texture and lower frequency features of images.

In addition, the fractional Brownian motion model has shown great success in simulating and parameterizing a wide variety of natural phenomena. In particular, the model provides a mathematical method for the analysis and representation of scale-invariant random textures \cite{tiedong2008method,lin2015alveolar,jiang2010forest,stewart1993fractional,you1997fractional,hoefer1993segmentation}. Moreover, fractional order Darwinian particle swarm optimization is proposed, where fractional derivative is used to control the convergence rate of particles. In this method, the problem of n-level thresholding is considered as an optimization problem to search the thresholds that maximize the between-class variance \cite{ghamisi2014multilevel}.
Fig.3.6 compares AC model, RSF model with their fractional-order methods, showing that the fractional-order method is superior than the integer order method, especially in the concave zone.

\begin{figure}[htbp]
  \centering
   \begin{subfigure}[b]{0.33\textwidth}
   \centering
        \includegraphics[width=1.5in]{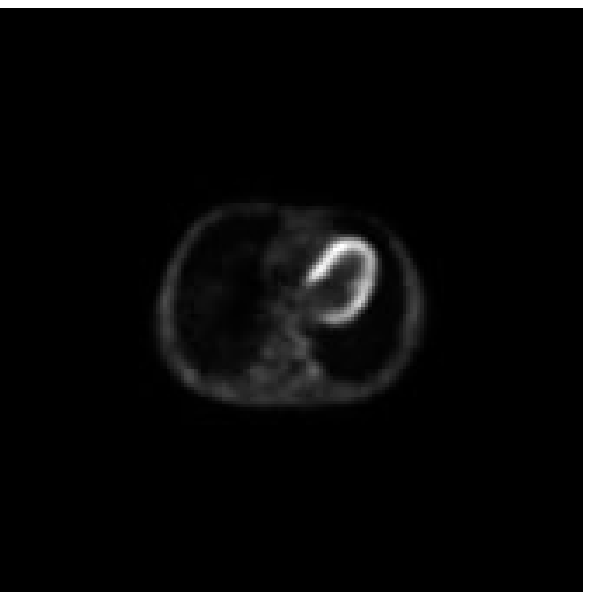}
        \caption{{\tt Original image}}
    \end{subfigure}%
   \begin{subfigure}[b]{0.33\textwidth}
   \centering
        \includegraphics[width=1.5in]{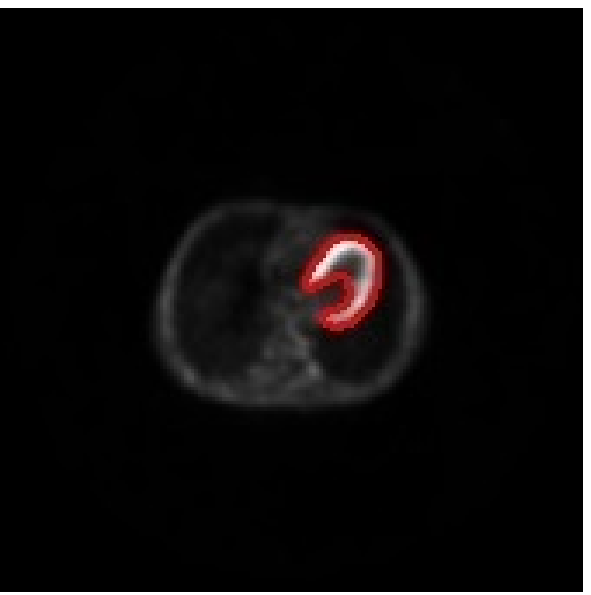}
        \caption{{\tt AC }}
   \end{subfigure}%
   \begin{subfigure}[b]{0.33\textwidth}
   \centering
        \includegraphics[width=1.5in]{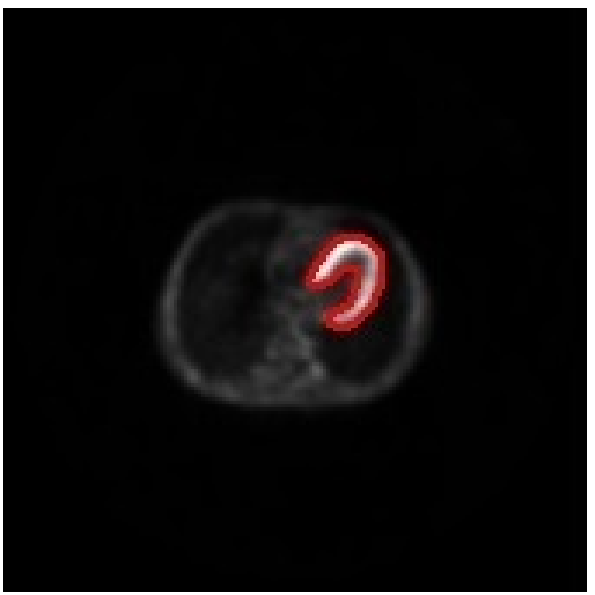}
        \caption{{\tt Fractional AC }}
   \end{subfigure}%
   \\
   \begin{subfigure}[b]{0.33\textwidth}
   \centering
        \includegraphics[width=1.5in]{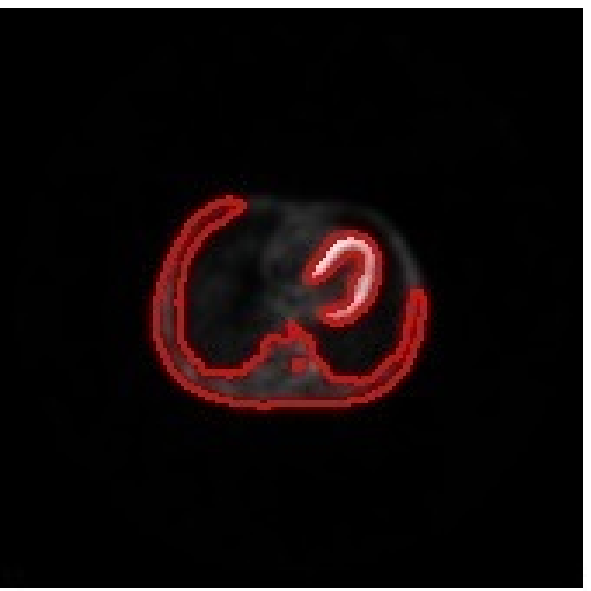}
        \caption{{\tt RSF }}
   \end{subfigure}%
   \begin{subfigure}[b]{0.33\textwidth}
   \centering
        \includegraphics[width=1.5in]{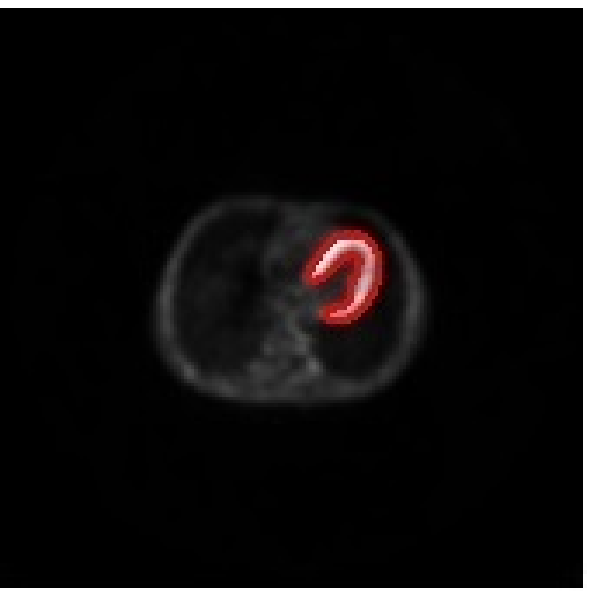}
        \caption{{\tt Fractional RSF }}
   \end{subfigure}%
   \bigskip

   Fig. 3.6: Fractional-order methods and other methods comparison result
  \label{fig:segmen-1}
\end{figure}

\subsection{Image Registration}
Nowadays, image registration has been applied in different fields, such as medicine, remote sensing and machine vision. Fractional-order derivative image registration methods have been proved many advantages to common methods. These algorithms can be classified according to the transformation models. The first category of transformation models refers to spatial transformations, which include rotation, scaling and translation. Spatial transformations cannot model local geometric differences between images because they are global in nature. To deal with this problem, matching key points with the fractional order SIFT method are used \cite{guimei2015reserch,pan2009adaptable,iwai2008security} (see fig.3.7). The second category of transformations enables non-rigid transformations. These transformations allow to locally register the template image to the reference image using fractional Fourier transform \cite{pan2009adaptable,niu2015image,zhang2013medical}. An adaptable multi-layer fractional Fourier transform approach for image registration is proposed in \cite{pan2009adaptable}. It generates lower interpolation errors in both polar and log-polar Fourier transform, and obtains higher accuracy with almost the same computing complexity as the pseudo-polar FFT. Niu et al. \cite{niu2015image} combine the fractional Fourier transform and conventional phase correlation technique for image registration which has been proved better noise immunity than FFT-based method. Zhang et al. \cite{zhang2013medical} compare two types fractional Fourier transforms techniques (discrete fractional Fourier transform and Shin's fractionalization of Fourier transform) in medical image registration. There are some other fractional Fourier transform methods that are used in images registration \cite{sharma2006image,li2012multilayer}. On the other hand, fractional-order based image registration algorithms can also be classified according to the variational model \cite{zhang2015variational} and differential equations \cite{melbourne2012using,garvey2013nonrigid}. Variational model approaches register images by minimizing the similarity measures, an index to quantify the degree of similarity between intensity patterns within two images. Zhang et al. \cite{zhang2015variational} use total fractional-order variation model to design variation regularization in non-rigid image registration. The model demonstrates substantial improvements in accuracy and robustness over the conventional image registration approaches. Recently, there have been several works about the image registration involving fractional order differential equations. Melbourne et al. \cite{melbourne2012using} design new gradients of image intensities using fractional differentiation and register image gradients directly to enhance image registration performance. Garvey et al. in \cite{garvey2013nonrigid} propose a non-rigid registration algorithm by directly and rapidly solving a discretized fractional PDE modeling super-diffusive process in non-rigid image registration. Experiments indicate that this algorithm has lower average deformation errors than standard diffusion-based registration in the registration.
Fig.3.7 compares fractional method of different $\alpha $ with original SIFT method. We can see that the fractional-order method extract more key points than the original SIFT method.
\begin{figure}[htbp]
  \centering
   \begin{subfigure}[b]{0.5\textwidth}
   \centering
        \includegraphics[width=2.3in]{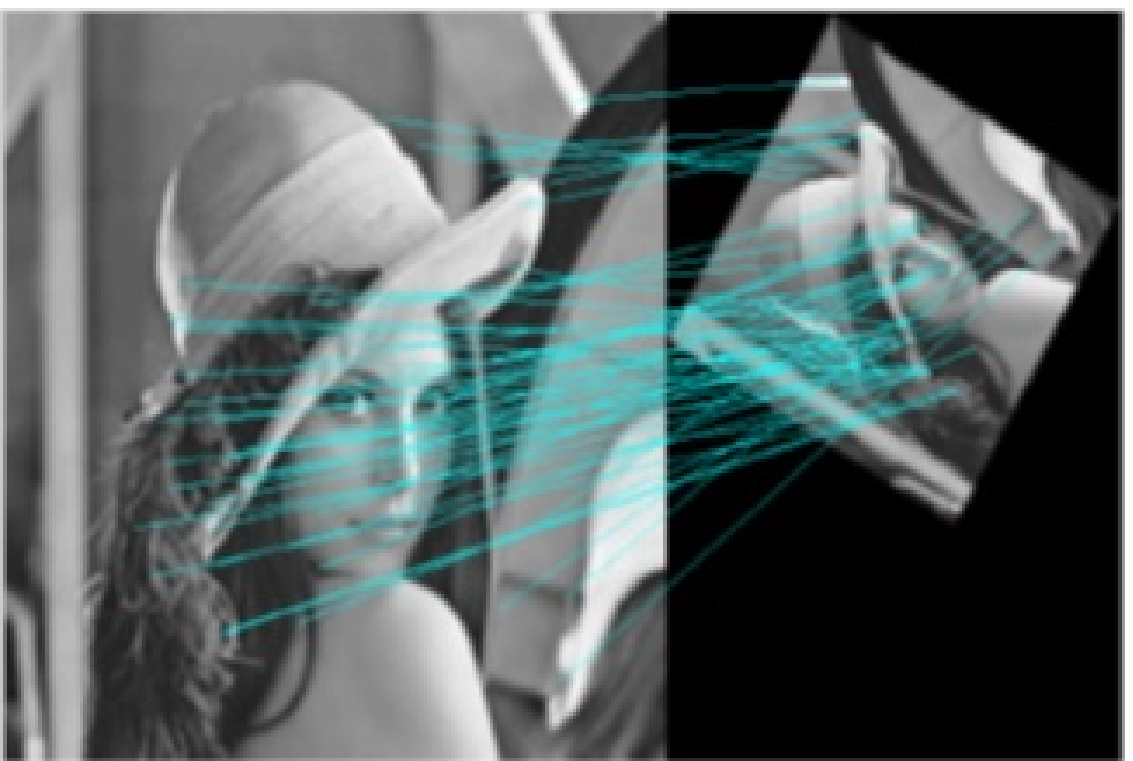}
        \caption{{\tt Original SIFT Algorithm}}
    \end{subfigure}%
   \begin{subfigure}[b]{0.5\textwidth}
   \centering
        \includegraphics[width=2.3in]{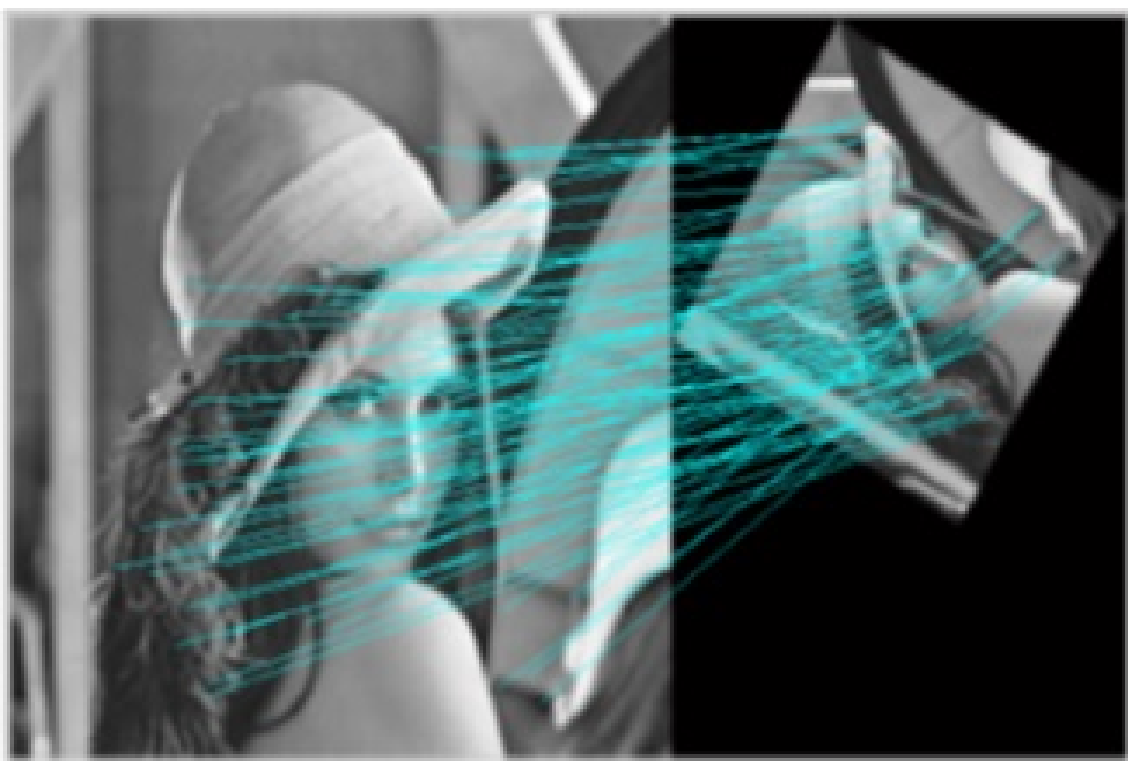}
        \caption{{\tt Fractional SIFT method $\alpha $ = 0.3}}
   \end{subfigure}%
   \\
     \centering
   \begin{subfigure}[b]{0.5\textwidth}
   \centering
        \includegraphics[width=2.3in]{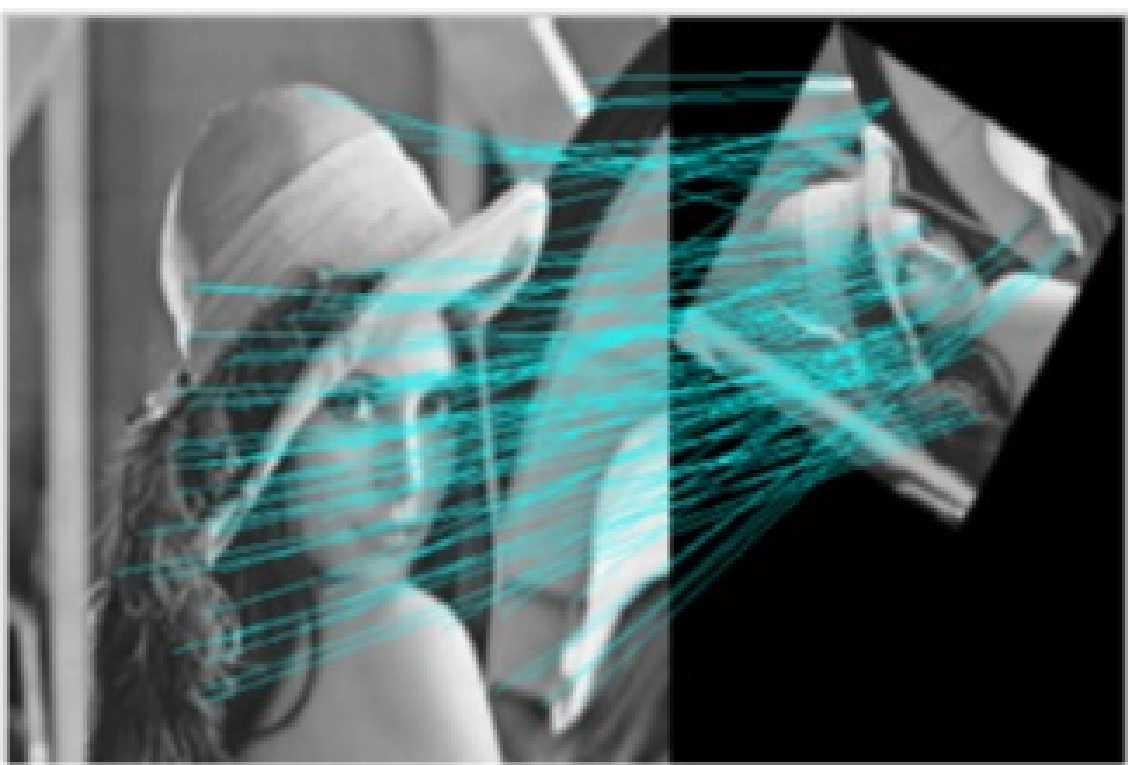}
        \caption{{\tt Fractional SIFT method $\alpha $ = 0.6}}
    \end{subfigure}%
   \begin{subfigure}[b]{0.5\textwidth}
   \centering
        \includegraphics[width=2.3in]{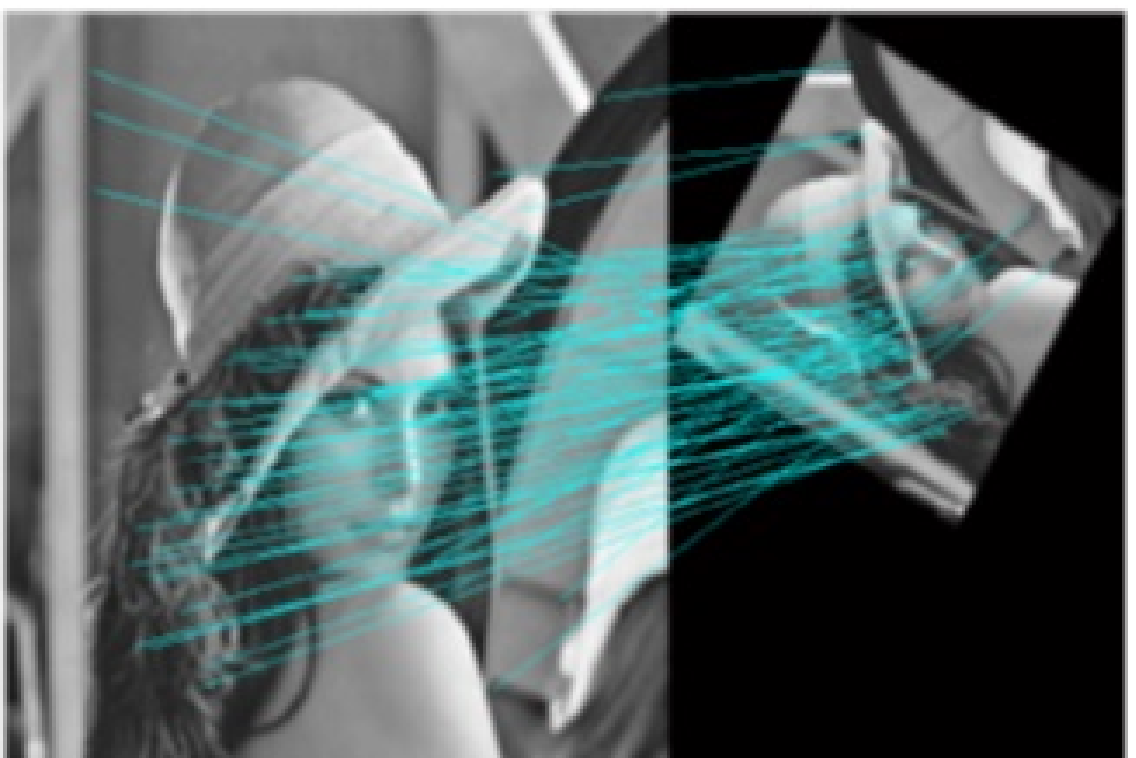}
        \caption{{\tt Fractional SIFT method $\alpha $ = 0.9}}
   \end{subfigure}%
   \label{fig:Registration}
   \\
    \bigskip

   Fig. 3.7: Comparison of different fractional $\alpha $'s method with original SIFT method \cite{guimei2015reserch}
    \label{fig:Registration}
\end{figure}

\subsection{Image Recognition}

Image recognition is an important research topic in pattern recognition. Fractional-order based method is used to reduce the dimensions of the image features for image recognition. It decomposes the image into a composition of a set of base images by singular value decomposition, and the top singular values are sensitive to great image variations, e.g., lighting, and occlusion. For instance, fractional-order embedding canonical correlation analysis \cite{yuan2014fractional} is proposed to reduce the dimensionality of multi-view data for classification tasks. Li et al. use fractional power polynomial model to reduce the dimensions of the textural features for image recognition \cite{li2012gabor}. Fractional low-order independent component analysis algorithm (FLOD-ICA) is constructed by Chen et al. to solve the partial occlusion face recognition problem \cite{chen2014fractional}. In addition, Kumar et al. design the joint fractional Fourier transform (JFRT) correlators with digital Fresnel holography technique for 3D object recognition \cite{kumar2015three}. Particularly, the JFRT may be used to detect a specific object appearing at a certain location because it is of shift-variance.
\subsection{Image Fusion}

Image fusion is the process of combining multiple images into a single image to increase the information content of the processed image. The researchers have developed various image fusion methods in the spatial domain and Fourier domain with different fusion rules like pixel averaging, maximum value selection, weighted average, region energy, region variance and so on. In these methods, fractional-order based fusion schemes provide additional degree of freedom in optimizing fusion quality. Fractional Fourier transform (FRFT) is the most famous method generalized from the FFT to analyze the signal in intermediate domains. In \cite{wang2013novel}, FRFT and non-subsampled contourlet transform (NSCT) are combined to exploit the local feature representation capability of NSCT and intermediate time-frequency representation capability of FRFT. Fusion method based on image decomposition using self-fractional Fourier functions (SFFF) is reported in \cite{sharma2014image,sharma2014hybrid}, where fusion quality of images is optimized by changing number of decomposition levels and by using some transform before SFFF decomposition. In addition, fusion method based on multi-parameter fractional random transform is proposed in \cite{lang2014novel,lang2015image}, according to which images are converted into the discrete multi-parameter fractional random transform domains respectively. The high amplitudes carry the spectral information and the low amplitudes just carry the spatial detail. Moreover, image fusion algorithm based on the fractional wavelet transform is used in \cite{tian2012new,xu2016medical,shkvarko2016radar}. It decomposes two source images to different frequency bands with fractional wavelet transform, then takes the largest absolute value fusion rules to gain the fusion coefficients, finally obtains the fusion image after fractional wavelet inverse transform. The simulation results prove that the performance of the fused image based on fractional wavelet transform is better than that based on conventional wavelet transform and fractional Fourier transform under the same condition.

Recently, image fusion based on fractional differential and variational model is proposed in \cite{li2016fractional}. The variational super-resolution and the image fusion method are combined to produce a novel variational model for image fusion and super-resolution in pixel level. By minimizing the energy function, the important structure information of the source images can be extracted and transferred into the fused image. The numerical results indicate that the proposed method is feasible and effective.

\subsection{Image Encryption}

 With the development of network and communication technology, image encryption plays an important role in information security. Usually, there are two methods to protect the encrypt image from brute-force attack for the key. One is to provide large enough key space to make such a search infeasible, the other method is to substitute the linear encryption method with the nonlinear encryption method for the higher security level. Fractional-order based methods have unique advantages in nonlinear control and key space extension and have been widely used in image encryption \cite{lima2014image,tao2010image,yoshimura2008new,elhoseny2016effect,zhao2016security}. One of the classical optical image encryption methods is double random phase encoding (DRPE). It is composed of two statistically independent masks of random phases where the first one is used in the spatial domain and the second one is applied as encryption key in the Fourier domain. With one fractional order parameter as an additional key, fractional Fourier transform (FrFT) has been used in DRPE in different ways to extend the key space. For example, a generalization of FrFT, multiple-parameter discrete fractional Fourier transform (MPDFrFT), can significantly increase DRPE key space by using multiple fractional orders\cite{pei2006multiple}. Upon the appearance of these techniques, hybrid opto-digital encryption methods have got a lot of attention \cite{lang2015color,sui2015double,ran2015image,zhou2015double,cui2013image,lang2010image}. In \cite{sui2015double}, Sui et al. generated digitally the masks of the DRPE using chaos maps, which are well-known of good cryptographic properties of high sensitivity to the initial parameters. In \cite{zhou2015double}, opto-digital encryption schemes have been proposed to enhance classical DRPE by applying digital random techniques. In \cite{lang2010image}, Lang et al. designed a permutation function in the MPDFrFT domains, which provides parameters of fractional orders and the logistic maps as the secret key. However, application of either Fourier transform(FT) or FrFT directly in classical DRPE is not safe in the aspects of security. It is easily attacked because the entire DRPE method is a linear transformation intrinsically \cite{qin2009vulnerability}. Therefore non-linear DRPE \cite{azoug2016non} is proposed to increase security for encryption of digital images, where the second mask is replaced with a chaotic permutation and Fourier transform is substituted with MPDFrFT. In addition, enhanced version of encryption scheme \cite{ahmad2015enhanced} is proposed to improve its attack defensibleness, plaintext sensitivity and inherent encryption strength.  Multiple-image encryption based on the cascaded fractional Fourier transform \cite{li2015asymmetric} is proposed. Each original image is directly separated into two phase masks. One is used as ciphertext image; the other is used as the encryption keys. The method has high resistance to various potential attacks, including the chosen-plaintext attack. Other methods include fractional matrix, fractional Mellin transform and multiple images encryotion\cite{li2015modified}.

\subsection{Image Compression}

The objective of image compression is to reduce irrelevance and redundancy of the image data meanwhile maintaining a high visual quality of decompressed images. The principle of image compression is to obtain the decomposition coefficients through specified basic function and then use the coefficient to express the image. Among these methods, fractional-order Fourier transform method has superior features owing to its extra degree of freedom, which provides extremely good compression rate, and allows to control degradation of image quality. It divides an image into non-overlapped $n\times n$ sub images firstly, and then calculates the decomposition coefficients in the frequency domain. Finally, these coefficients are normalized according to the selected cutoff value as image representation. This method provides better compression rate by tuning the fractional order in FrFT \cite{singh2009image,kumar2013digital}. By selecting different basic functions, different image decomposition coefficients will be obtained to represent the compressed image. For example, in \cite{yetik2001image} an image is approximately represented as the coefficients of the minimum mean square with FrFT method. An image is approximately represented as the coefficients of fractional Fourier transform with a wavelet algorithm and Huffman encoder \cite{kumari2013secure}. Recently, image compression decomposition coefficients obtained from nonlinear fractional Mellin transform (FrMT) \cite{zhou2015image} is proposed. The method provides a convenient way to compress image. It measure the original image though measurement matrices, and then takes nonlinear FrMT to decrease the data volume of the decomposition image. In this process, the measurements are performed in two directions and the measurement matrices are designed as partial Hadamard matrices. The simulation results show its good security to resist common attacks due to nonlinearity and high compression ratio.

\subsection{Image Restoration}
  The encoded block stream data might be damaged or lost though the unsafe network protocol transmission, especially in the user datagram protocol (UDP) transmission. Also the focus might be lost during image capture, especially in motion object image capture. Image restoration has got widely used when some part of an image is missing, incomplete, degraded or distorted. One problem of 2-D image restoration is obstacle background estimation and lost block recovery. Another one is recovery of blurred defocused image and degraded image. For the first problem, Serbes et al. propose an image recovery approach by alternating projections onto convex sets in fractional Fourier domains \cite{serbes2010optimum}. Here the optimum fractional Fourier domain is obtained by minimizing the bandwidth of fractional Fourier domain order, and alternating projections are used for image recovery problems. Yang et al. use defocused imaging model based on fractional Fourier transform (FrFT) to illustrate the blurring phenomena and then inverse FrFT is implemented to resharpen image \cite{yang2007blurred}. In order to improve imaging clarity, retina-filtering model is used, whose parameters were determined by fractional diffusion equation \cite{paek2013retina}. For the second problem, total variation model \cite{chen2014automatic,chen2016fractional,kumar2016fractional} and lots of filters in fractional Fourier domain are used, such as repeated filtering in consecutive fractional Fourier domains for image restoration \cite{erden1999repeated}, image restoration based on the fraction Fourier transform \cite{yan2001image}, optimal image restoration with the fractional Fourier transform \cite{kutay1998optimal} and fractional nonlinear anisotropic diffusion method \cite{yin2015fractional}.

\section{Conclusion}
\label{sec:conclusion}

This paper introduces the fractional-order definitions, discretization methods and related Matlab toolboxes. Moreover, an extensive collection of fractional-order based image processing methods are exhibited to introduce the methodology of fractional-order image processing, as well as some experimental data, with an emphasis on fractional order tuning.
\section*{Acknowledgements}
The work was partially supported by the Liaoning Province Education Project No. L2014082 and China Scholarship Council.

\bibliography{FCAA_Image_Processing}

\begin{thebibliography}{100}
\providecommand{\url}[1]{#1}
\csname url@samestyle\endcsname
\providecommand{\newblock}{\relax}
\providecommand{\bibinfo}[2]{#2}
\providecommand{\BIBentrySTDinterwordspacing}{\spaceskip=0pt\relax}
\providecommand{\BIBentryALTinterwordstretchfactor}{4}
\providecommand{\BIBentryALTinterwordspacing}{\spaceskip=\fontdimen2\font plus
\BIBentryALTinterwordstretchfactor\fontdimen3\font minus
  \fontdimen4\font\relax}
\providecommand{\BIBforeignlanguage}[2]{{%
\expandafter\ifx\csname l@#1\endcsname\relax
\typeout{** WARNING: IEEEtran.bst: No hyphenation pattern has been}%
\typeout{** loaded for the language `#1'. Using the pattern for}%
\typeout{** the default language instead.}%
\else
\language=\csname l@#1\endcsname
\fi
#2}}
\providecommand{\BIBdecl}{\relax}
\BIBdecl

\bibitem{petravs2011fractional}
I.~Petr{\'a}{\v{s}}, ``Fractional derivatives, fractional integrals, and
  fractional differential equations in matlab,'' \emph{Engineering Education
  and Research Using MATLAB, InTech, kap}, vol.~10, pp. 239--264, 2011.

\bibitem{chen2002discretization}
Y.~Q. Chen and K.~L. Moore, ``Discretization schemes for fractional-order
  differentiators and integrators,'' \emph{Circuits and Systems I: Fundamental
  Theory and Applications, IEEE Transactions on}, vol.~49, no.~3, pp. 363--367,
  2002.

\bibitem{podlubny2000matrix}
I.~Podlubny, ``Matrix approach to discrete fractional calculus,''
  \emph{Fractional Calculus and Applied Analysis}, vol.~3, no.~4, pp. 359--386,
  2000.

\bibitem{oustaloup2000frequency}
A.~Oustaloup, F.~Levron, B.~Mathieu, and F.~M. Nanot, ``Frequency-band complex
  noninteger differentiator: characterization and synthesis,'' \emph{Circuits
  and Systems I: Fundamental Theory and Applications, IEEE Transactions on},
  vol.~47, no.~1, pp. 25--39, 2000.

\bibitem{magin2006fractional}
R.~L. Magin, \emph{Fractional Calculus in Bioengineering}.\hskip 1em plus 0.5em
  minus 0.4em\relax Begell House Redding, 2006.

\bibitem{magin2008fractional}
R.~Magin, X.~Feng, and D.~Baleanu, ``Fractional calculus in {NMR},'' \emph{IFAC
  Proceedings Volumes}, vol.~41, no.~2, pp. 9613--9618, 2008.

\bibitem{magin2008modeling}
R.~Magin and M.~Ovadia, ``Modeling the cardiac tissue electrode interface using
  fractional calculus,'' \emph{Journal of Vibration and Control}, vol.~14, no.
  9-10, pp. 1431--1442, 2008.

\bibitem{magin2008anomalous}
R.~L. Magin, O.~Abdullah, D.~Baleanu, and X.~J. Zhou, ``Anomalous diffusion
  expressed through fractional order differential operators in the
  bloch--torrey equation,'' \emph{Journal of Magnetic Resonance}, vol. 190,
  no.~2, pp. 255--270, 2008.

\bibitem{dingyu2006control}
X.~Dingyu, \emph{Control System Computer Aided Design-MATLAB Language and
  Application}.\hskip 1em plus 0.5em minus 0.4em\relax Tsinghua University
  Press, 2006.

\bibitem{Jonathan2014Fractional}
Jonathan, ``Fractional derivative,'' [Online] (MatlabCentral)
  \url{http://www.mathworks.com/matlabcentral/fileexchange/45982}, 2014.

\bibitem{podlubny2009matrixartical}
I.~Podlubny, A.~Chechkin, T.~Skovranek, Y.~Chen, and B.~M.~V. Jara, ``Matrix
  approach to discrete fractional calculus ii: Partial fractional differential
  equations,'' \emph{Journal of Computational Physics}, vol. 228, no.~8, pp.
  3137--3153, 2009.

\bibitem{Podlubny2009Matrix}
I.~Podlubny, ``Matrix approach to discretization of odes and pdes of arbitrary
  real order,'' [Online] (MatlabCentral)
  \url{http://www.mathworks.com/matlabcentral/fileexchange/22071}, 2009.

\bibitem{li2013image}
X.~M. Li, ``Image enhancement in the fractional {Fourier} domain,'' in
  \emph{Proceedings of 6th International Congress on Image and Signal
  Processing {(CISP)}}, vol.~1.\hskip 1em plus 0.5em minus 0.4em\relax IEEE,
  2013, pp. 299--303.

\bibitem{roy2015fractional}
S.~Roy, P.~Shivakumara, H.~A. Jalab, R.~W. Ibrahim, U.~Pal, and T.~Lu,
  ``Fractional poisson enhancement model for text detection and recognition in
  video frames,'' \emph{Pattern Recognition}, 2015.

\bibitem{li2016fractional}
H.~Li, Z.~Yu, and C.~Mao, ``Fractional differential and variational method for
  image fusion and super-resolution,'' \emph{Neurocomputing}, vol. 171, pp.
  138--148, 2016.

\bibitem{pu2010fractional}
Y.~Pu, J.~Zhou, and X.~Yuan, ``Fractional differential mask: a fractional
  differential-based approach for multiscale texture enhancement,'' \emph{Image
  Processing, IEEE Transactions on}, vol.~19, no.~2, pp. 491--511, 2010.

\bibitem{chen20121}
D.~Chen, Y.~Chen, and D.~Xue, ``{1-D} and {2-D} digital fractional-order
  savitzky--golay differentiator,'' \emph{Signal, Image and Video Processing},
  vol.~6, no.~3, pp. 503--511, 2012.

\bibitem{li2015adaptive}
B.~Li and W.~Xie, ``Adaptive fractional differential approach and its
  application to medical image enhancement,'' \emph{Computers and Electrical
  Engineering}, 2015.

\bibitem{li2015image}
------, ``Image denoising and enhancement based on adaptive fractional calculus
  of small probability strategy,'' \emph{Neurocomputing}, 2015.

\bibitem{hu2015adaptive}
F.~Hu, S.~Si, H.~San~Wong, B.~Fu, M.~Si, and H.~Luo, ``An adaptive approach for
  texture enhancement based on a fractional differential operator with
  non-integer step and order,'' \emph{Neurocomputing}, vol. 158, pp. 295--306,
  2015.

\bibitem{si2014texture}
M.~Si, L.~Fang, F.~Hu, and S.~Si, ``Texture enhancement algorithm based on
  fractional differential mask of adaptive non-integral step,'' in
  \emph{Proceedings of 7th International Congress on Image and Signal
  Processing {(CISP)}}.\hskip 1em plus 0.5em minus 0.4em\relax IEEE, 2014, pp.
  179--183.

\bibitem{che2012fractional}
J.~Che, Y.-S. Shi, Y.~Xiang, and Y.-T. Ma, ``The fractional differential
  enhancement of image texture features and its parallel processing
  optimization,'' in \emph{Proceedings of 5th International Congress on Image
  and Signal Processing {(CISP)}}.\hskip 1em plus 0.5em minus 0.4em\relax IEEE,
  2012, pp. 330--333.

\bibitem{li2016image}
B.~Li and W.~Xie, ``Image denoising and enhancement based on adaptive
  fractional calculus of small probability strategy,'' \emph{Neurocomputing},
  vol. 175, pp. 704--714, 2016.

\bibitem{he2015improved}
N.~He, J.-B. Wang, L.-L. Zhang, and K.~Lu, ``An improved fractional-order
  differentiation model for image denoising,'' \emph{Signal Processing}, vol.
  112, pp. 180--188, 2015.

\bibitem{jalab2015fractional}
H.~A. Jalab and R.~W. Ibrahim, ``Fractional alexander polynomials for image
  denoising,'' \emph{Signal Processing}, vol. 107, pp. 340--354, 2015.

\bibitem{che2013image}
J.~Che, Q.~Guan, and X.~Wang, ``Image denoising based on adaptive fractional
  partial differential equations,'' in \emph{Proceedings of 6th International
  Congress on Image and Signal Processing {(CISP)}}, vol.~1.\hskip 1em plus
  0.5em minus 0.4em\relax IEEE, 2013, pp. 288--292.

\bibitem{jun2011class}
Z.~Jun and W.~Zhihui, ``A class of fractional-order multi-scale variational
  models and alternating projection algorithm for image denoising,''
  \emph{Applied Mathematical Modelling}, vol.~35, no.~5, pp. 2516--2528, 2011.

\bibitem{tian2015fractional}
D.~Tian, D.~Xue, and D.~Wang, ``A fractional-order adaptive regularization
  primal--dual algorithm for image denoising,'' \emph{Information Sciences},
  vol. 296, pp. 147--159, 2015.

\bibitem{hu2012new}
X.~Hu and Y.~Li, ``A new variational model for image denoising based on
  fractional-order derivative,'' in \emph{Proceedings of International
  Conference on Systems and Informatics {(ICSAI2012)}}, 2012.

\bibitem{dong2013variational}
F.~Dong, ``A variational framework for image denoising based on
  fractional-order derivatives,'' in \emph{Proceedings of Ninth International
  Conference on Natural Computation {(ICNC)}}.\hskip 1em plus 0.5em minus
  0.4em\relax IEEE, 2013, pp. 1283--1288.

\bibitem{chen2015fractional}
D.~Chen, Y.~Chen, and D.~Xue, ``Fractional-order total variation image
  denoising based on proximity algorithm,'' \emph{Applied Mathematics and
  Computation}, vol. 257, pp. 537--545, 2015.

\bibitem{chen2013fractional}
D.~Chen, S.~Sun, C.~Zhang, Y.~Chen, and D.~Xue, ``Fractional-order {TV-L2}
  model for image denoising,'' \emph{Central European Journal of Physics},
  vol.~11, no.~10, pp. 1414--1422, 2013.

\bibitem{chen2016robust}
G.~Chen, J.~Zhang, D.~Li, and H.~Chen, ``Robust kronecker product video
  denoising based on fractional-order total variation model,'' \emph{Signal
  Processing}, vol. 119, pp. 1--20, 2016.

\bibitem{zhang2015total}
J.~Zhang and K.~Chen, ``A total fractional-order variation model for image
  restoration with nonhomogeneous boundary conditions and its numerical
  solution,'' \emph{SIAM Journal on Imaging Sciences}, vol.~8, no.~4, pp.
  2487--2518, 2015.

\bibitem{brito2010multigrid}
C.~Brito-Loeza and K.~Chen, ``Multigrid algorithm for high order denoising,''
  \emph{SIAM Journal on Imaging Sciences}, vol.~3, no.~3, pp. 363--389, 2010.

\bibitem{chan2010multilevel}
R.~H. Chan and K.~Chen, ``A multilevel algorithm for simultaneously denoising
  and deblurring images,'' \emph{SIAM Journal on Scientific Computing},
  vol.~32, no.~2, pp. 1043--1063, 2010.

\bibitem{chan2006optimization}
T.~F. Chan and K.~Chen, ``An optimization-based multilevel algorithm for total
  variation image denoising,'' \emph{Multiscale Modeling and Simulation},
  vol.~5, no.~2, pp. 615--645, 2006.

\bibitem{zhang2012iterative}
J.~Zhang, K.~Chen, and B.~Yu, ``An iterative lagrange multiplier method for
  constrained total-variation-based image denoising,'' \emph{SIAM Journal on
  Numerical Analysis}, vol.~50, no.~3, pp. 983--1003, 2012.

\bibitem{zachevsky2016statistics}
I.~Zachevsky and Y.~Y.~J. Zeevi, ``Statistics of natural stochastic textures
  and their application in image denoising,'' \emph{IEEE Transactions on Image
  Processing}, vol.~25, no.~5, pp. 2130--2145, 2016.

\bibitem{zhang2014fractional}
W.~Zhang, J.~Li, and Y.~Yang, ``A fractional diffusion-wave equation with
  non-local regularization for image denoising,'' \emph{Signal Processing},
  vol. 103, pp. 6--15, 2014.

\bibitem{bai2007fractional}
J.~Bai and X.~Feng, ``Fractional-order anisotropic diffusion for image
  denoising,'' \emph{Image Processing, IEEE Transactions on}, vol.~16, no.~10,
  pp. 2492--2502, 2007.

\bibitem{zhang2015spatial}
W.~Zhang, J.~Li, and Y.~Yang, ``Spatial fractional telegraph equation for image
  structure preserving denoising,'' \emph{Signal Processing}, vol. 107, pp.
  368--377, 2015.

\bibitem{rudin1992nonlinear}
L.~I. Rudin, S.~Osher, and E.~Fatemi, ``Nonlinear total variation based noise
  removal algorithms,'' \emph{Physica D: Nonlinear Phenomena}, vol.~60, no.~1,
  pp. 259--268, 1992.

\bibitem{gilles2010properties}
J.~Gilles and Y.~Meyer, ``Properties of {BV- G} structures+ textures
  decomposition models. application to road detection in satellite images,''
  \emph{IEEE Transactions on Image Processing}, vol.~19, no.~11, pp.
  2793--2800, 2010.

\bibitem{pu2006fractional}
Y.~Pu, ``Fractional calculus approach to texture of digital image,'' in
  \emph{Proceedings of 8th international Conference on Signal Processing},
  2006.

\bibitem{yi2007fractional}
Y.~Fei, ``Fractional differential analysis for texture of digital image,''
  \emph{Journal of Algorithms and Computational Technology}, vol.~1, no.~3, pp.
  357--380, 2007.

\bibitem{tian2014fractional}
D.~Tian, J.~Wu, and Y.~Yang, ``A fractional-order edge detection operator for
  medical image structure feature extraction,'' in \emph{Proceedings of 26th
  Chinese Control and Decision Conference {(CCDC)}}.\hskip 1em plus 0.5em minus
  0.4em\relax IEEE, 2014, pp. 5173--5176.

\bibitem{yang2010novel}
H.~Yang, Y.~Ye, D.~Wang, and B.~Jiang, ``A novel fractional-order signal
  processing based edge detection method,'' in \emph{Proceedings of 11th
  International Conference on Control Automation Robotics and Vision}, 2010,
  pp. 1122--1127.

\bibitem{mathieu2003fractional}
B.~Mathieu, P.~Melchior, A.~Oustaloup, and C.~Ceyral, ``Fractional
  differentiation for edge detection,'' \emph{Signal Processing}, vol.~83,
  no.~11, pp. 2421--2432, 2003.

\bibitem{gao2011edge}
C.~Gao, J.~Zhou, J.~Hu, and F.~Lang, ``Edge detection of colour image based on
  quaternion fractional differential,'' \emph{IET Image Processing}, vol.~5,
  no.~3, pp. 261--272, 2011.

\bibitem{pan2013novel}
X.~Pan, Y.~Ye, J.~Wang, and X.~Gao, ``Novel fractional-order calculus masks and
  compound derivatives with applications to edge detection,'' in
  \emph{Proceedings of 6th International Congress on Image and Signal
  Processing {(CISP)}}, vol.~1.\hskip 1em plus 0.5em minus 0.4em\relax IEEE,
  2013, pp. 309--314.

\bibitem{gao2014edge}
C.~Gao, J.~Zhou, and W.-h. Zhang, ``Edge detection based on the newton
  interpolation's fractional differentiation.'' \emph{Int. Arab J. Inf.
  Technol.}, vol.~11, no.~3, pp. 223--228, 2014.

\bibitem{chen2012Fractionalrobust}
Y.~C. Dali~Chen, DingyuXue, ``Fractional differentiation-based approach for
  robust image edge detection,'' in \emph{Proceedings of Conference on
  Fractional Derivative and Applications}.\hskip 1em plus 0.5em minus
  0.4em\relax FDA, 2012, pp. 322--327.

\bibitem{tian2013fractional}
D.~Tian, D.~Xue, D.~Chen, and S.~Sun, ``A fractional-order regulatory {CV}
  model for brain {MR} image segmentation,'' in \emph{Proceedings of 25th
  Chinese Control and Decision Conference {(CCDC)}}.\hskip 1em plus 0.5em minus
  0.4em\relax IEEE, 2013, pp. 37--40.

\bibitem{ren2015adaptive}
Z.~Ren, ``Adaptive active contour model driven by fractional order fitting
  energy,'' \emph{Signal Processing}, vol. 117, pp. 138--150, 2015.

\bibitem{zhang2014local}
J.~Zhang, K.~Chen, B.~Yu, and D.~A. Gould, ``A local information based
  variational model for selective image segmentation,'' \emph{Inverse Problems
  Imaging}, vol.~8, no.~1, pp. 293--320, 2014.

\bibitem{tiedong2008method}
Z.~Tiedong, W.~Lei, Q.~Zaibai, and L.~Yu, ``A method of underwater image
  segmentation based on discrete fractional brownian random field,'' in
  \emph{Proceedings of 3rd Conference on Industrial Electronics and
  Applications}.\hskip 1em plus 0.5em minus 0.4em\relax IEEE, 2008, pp.
  2507--2511.

\bibitem{lin2015alveolar}
P.~Lin, P.~Huang, P.~Huang, and H.~Hsu, ``Alveolar bone-loss area localization
  in periodontitis radiographs based on threshold segmentation with a hybrid
  feature fused of intensity and the {H-value} of fractional {Brownian} motion
  model,'' \emph{Computer methods and programs in biomedicine}, vol. 121,
  no.~3, pp. 117--126, 2015.

\bibitem{jiang2010forest}
A.~Jiang and L.~Liu, ``Forest fire image segmentation based on contourlet
  transform and fractional {Brownian} motion,'' in \emph{Proceedings of
  International Workshop on Chaos-Fractals Theories and Applications
  {(IWCFTA)}}.\hskip 1em plus 0.5em minus 0.4em\relax IEEE, 2010, pp. 425--429.

\bibitem{stewart1993fractional}
C.~V. Stewart, B.~Moghaddam, K.~J. Hintz, and L.~M. Novak, ``Fractional
  brownian motion models for synthetic aperture radar imagery scene
  segmentation,'' \emph{Proceedings of the IEEE}, vol.~81, no.~10, pp.
  1511--1522, 1993.

\bibitem{you1997fractional}
J.~You, S.~Hungnahally, and A.~Sattar, ``Fractional discrimination for texture
  image segmentation,'' in \emph{Proceedings of International Conference on
  Image Processing}, vol.~1.\hskip 1em plus 0.5em minus 0.4em\relax IEEE, 1997,
  pp. 220--223.

\bibitem{hoefer1993segmentation}
S.~Hoefer, F.~Heil, M.~Pandit, and R.~Kumaresan, ``Segmentation of textures
  with different roughness using the model of isotropic two-dimensional
  fractional {Brownian} motion,'' in \emph{Proceedings of International
  Conference on Acoustics, Speech, and Signal Processing}, vol.~5.\hskip 1em
  plus 0.5em minus 0.4em\relax IEEE, 1993, pp. 53--56.

\bibitem{ghamisi2014multilevel}
P.~Ghamisi, M.~S. Couceiro, F.~M. Martins, and J.~Atli~Benediktsson,
  ``Multilevel image segmentation based on fractional-order {Darwinian}
  particle swarm optimization,'' \emph{Geoscience and Remote Sensing, IEEE
  Transactions on}, vol.~52, no.~5, pp. 2382--2394, 2014.

\bibitem{guimei2015reserch}
Y.~C. Guimei~Zhang, Binbin~Chen, ``Research on image matching combining on
  fractional differential with scale invariant feature transform,'' in
  \emph{Proceedings of Conference on International Design Engineering Technical
  and Computers and Information in Engineering}.\hskip 1em plus 0.5em minus
  0.4em\relax ASME, 2015, pp. 10--15.

\bibitem{pan2009adaptable}
W.~Pan, K.~Qin, and Y.~Chen, ``An adaptable-multilayer fractional {Fourier}
  transform approach for image registration,'' \emph{Pattern Analysis and
  Machine Intelligence, IEEE Transactions on}, vol.~31, no.~3, pp. 400--414,
  2009.

\bibitem{iwai2008security}
R.~Iwai and H.~Yoshimura, ``Security of registration data of fingerprint image
  with a server by use of the fractional {Fourier} transform,'' in
  \emph{Proceedings of 9th International Conference on Signal
  Processing}.\hskip 1em plus 0.5em minus 0.4em\relax IEEE, 2008, pp.
  2070--2073.

\bibitem{niu2015image}
H.~Niu, E.~Chen, L.~Qi, and X.~Guo, ``Image registration based on fractional
  {Fourier} transform,'' \emph{Optik-International Journal for Light and
  Electron Optics}, vol. 126, no.~23, pp. 3889--3893, 2015.

\bibitem{zhang2013medical}
X.~Zhang, Y.~Shen, S.~Li, and H.~Zhang, ``Medical image registration in
  fractional {Fourier} transform domain,'' \emph{Optik-International Journal
  for Light and Electron Optics}, vol. 124, no.~12, pp. 1239--1242, 2013.

\bibitem{sharma2006image}
K.~K. Sharma and S.~D. Joshi, ``Image registration using fractional {Fourier}
  transform,'' in \emph{Proceedings of Asia Pacific Conference on Circuits and
  Systems}.\hskip 1em plus 0.5em minus 0.4em\relax IEEE, 2006, pp. 470--473.

\bibitem{li2012multilayer}
Z.~Li, J.~Yang, R.~Lan, and X.~Feng, ``Multilayer-pseudopolar fractional
  {Fourier} transform approach for image registration,'' in \emph{Proceedings
  of Eighth International Conference on Computational Intelligence and
  Security}.\hskip 1em plus 0.5em minus 0.4em\relax IEEE, 2012, pp. 323--327.

\bibitem{zhang2015variational}
J.~Zhang and K.~Chen, ``Variational image registration by a total
  fractional-order variation model,'' \emph{Journal of Computational Physics},
  vol. 293, pp. 442--461, 2015.

\bibitem{melbourne2012using}
A.~Melbourne, N.~Cahill, C.~Tanner, M.~Modat, D.~Hawkes, and S.~Ourselin,
  ``Using fractional gradient information in non-rigid image registration:
  application to breast {MRI},'' in \emph{Proceedings of Conference on SPIE
  Medical Imaging}.\hskip 1em plus 0.5em minus 0.4em\relax International
  Society for Optics and Photonics, 2012, pp. 83\,141Z--83\,141Z.

\bibitem{garvey2013nonrigid}
C.~C. Garvey, N.~D. Cahill, A.~Melbourne, C.~Tanner, S.~Ourselin, and D.~J.
  Hawkes, ``Nonrigid image registration with two-sided space-fractional partial
  differential equations,'' in \emph{Proceedings of 20th International
  Conference on Image Processing {(ICIP)}}.\hskip 1em plus 0.5em minus
  0.4em\relax IEEE, 2013, pp. 747--751.

\bibitem{yuan2014fractional}
Y.~Yuan, Q.~Sun, and H.~Ge, ``Fractional-order embedding canonical correlation
  analysis and its applications to multi-view dimensionality reduction and
  recognition,'' \emph{Pattern Recognition}, vol.~47, no.~3, pp. 1411--1424,
  2014.

\bibitem{li2012gabor}
J.~Li, ``Gabor filter based optical image recognition using fractional power
  polynomial model based common discriminant locality preserving projection
  with kernels,'' \emph{Optics and Lasers in Engineering}, vol.~50, no.~9, pp.
  1281--1286, 2012.

\bibitem{chen2014fractional}
X.~Chen, W.~Li, Z.~Liu, and Z.~Zhou, ``Fractional low-order independent
  component analysis for face recognition robust to partial occlusion,'' in
  \emph{Proceedings of International Symposium on Biometrics and Security
  Technologies {(ISBAST)}}.\hskip 1em plus 0.5em minus 0.4em\relax IEEE, 2014,
  pp. 1--5.

\bibitem{kumar2015three}
D.~Kumar and N.~K. Nishchal, ``Three-dimensional object recognition using joint
  fractional {Fourier} transform correlators with the help of digital fresnel
  holography,'' \emph{Optik-International Journal for Light and Electron
  Optics}, vol. 126, no.~20, pp. 2690--2695, 2015.

\bibitem{wang2013novel}
P.~Wang, H.~Tian, and W.~Zheng, ``A novel image fusion method based on
  frft-nsct,'' \emph{Mathematical Problems in Engineering}, vol. 2013, 2013.

\bibitem{sharma2014image}
K.~Sharma and M.~Sharma, ``Image fusion based on image decomposition using
  self-fractional {Fourier} functions,'' \emph{Signal, image and video
  processing}, vol.~8, no.~7, pp. 1335--1344, 2014.

\bibitem{sharma2014hybrid}
J.~Sharma, K.~Sharma, and V.~Sahula, ``Hybrid image fusion scheme using
  self-fractional {Fourier} functions and multivariate empirical mode
  decomposition,'' \emph{Signal Processing}, vol. 100, pp. 146--159, 2014.

\bibitem{lang2014novel}
J.~Lang and Z.~Hao, ``Novel image fusion method based on adaptive pulse coupled
  neural network and discrete multi-parameter fractional random transform,''
  \emph{Optics and Lasers in Engineering}, vol.~52, pp. 91--98, 2014.

\bibitem{lang2015image}
------, ``Image fusion method based on adaptive pulse coupled neural network in
  the discrete fractional random transform domain,'' \emph{Optik-International
  Journal for Light and Electron Optics}, vol. 126, no.~23, pp. 3644--3651,
  2015.

\bibitem{tian2012new}
H.~Tian, P.~Wang, and W.~Zheng, ``A new image fusion algorithm based on
  fractional wavelet transform,'' in \emph{Proceedings of 2nd International
  Conference on Computer Science and Network Technology {(ICCSNT)}}.\hskip 1em
  plus 0.5em minus 0.4em\relax IEEE, 2012, pp. 2175--2178.

\bibitem{xu2016medical}
X.~Xu, Y.~Wang, and S.~Chen, ``Medical image fusion using discrete fractional
  wavelet transform,'' \emph{Biomedical Signal Processing and Control},
  vol.~27, pp. 103--111, 2016.

\bibitem{shkvarko2016radar}
Y.~V. Shkvarko, J.~I. Ya{\~n}ez, J.~A. Amao, and G.~D.~M. del Campo,
  ``Radar/sar image resolution enhancement via unifying descriptive experiment
  design regularization and wavelet-domain processing,'' \emph{IEEE Geoscience
  and Remote Sensing Letters}, vol.~13, no.~2, pp. 152--156, 2016.

\bibitem{lima2014image}
J.~Lima and L.~Novaes, ``Image encryption based on the fractional {Fourier}
  transform over finite fields,'' \emph{Signal Processing}, vol.~94, pp.
  521--530, 2014.

\bibitem{tao2010image}
R.~Tao, X.-Y. Meng, and Y.~Wang, ``Image encryption with multiorders of
  fractional {Fourier} transforms,'' \emph{Information Forensics and Security,
  IEEE Transactions on}, vol.~5, no.~4, pp. 734--738, 2010.

\bibitem{yoshimura2008new}
H.~Yoshimura and R.~Iwai, ``New encryption method of {2D} image by use of the
  fractional {Fourier} transform,'' in \emph{Proceedings of 9th International
  Conference on Signal Processing}.\hskip 1em plus 0.5em minus 0.4em\relax
  IEEE, 2008, pp. 2182--2184.

\bibitem{elhoseny2016effect}
H.~M. Elhoseny, O.~S. Faragallah, H.~E. Ahmed, H.~B. Kazemian, H.~S. El-sayed,
  and F.~E.~A. El-Samie, ``The effect of fractional {Fourier} transform angle
  in encryption quality for digital images,'' \emph{Optik-International Journal
  for Light and Electron Optics}, vol. 127, no.~1, pp. 315--319, 2016.

\bibitem{zhao2016security}
T.~Zhao, Q.~Ran, L.~Yuan, Y.~Chi, and J.~Ma, ``Security of image encryption
  scheme based on multi-parameter fractional {F}ourier transform,''
  \emph{Optics Communications}, vol. 376, pp. 47--51, 2016.

\bibitem{pei2006multiple}
S.~Pei and W.~Hsue, ``The multiple-parameter discrete fractional {Fourier}
  transform,'' \emph{Signal Processing Letters, IEEE}, vol.~13, no.~6, pp.
  329--332, 2006.

\bibitem{lang2015color}
J.~Lang, ``Color image encryption based on color blend and chaos permutation in
  the reality-preserving multiple-parameter fractional {Fourier} transform
  domain,'' \emph{Optics Communications}, vol. 338, pp. 181--192, 2015.

\bibitem{sui2015double}
L.~Sui, K.~Duan, and J.~Liang, ``Double-image encryption based on discrete
  multiple-parameter fractional angular transform and two-coupled logistic
  maps,'' \emph{Optics Communications}, vol. 343, pp. 140--149, 2015.

\bibitem{ran2015image}
Q.~Ran, L.~Yuan, and T.~Zhao, ``Image encryption based on nonseparable
  fractional {Fourier} transform and chaotic map,'' \emph{Optics
  Communications}, vol. 348, pp. 43--49, 2015.

\bibitem{zhou2015double}
N.~Zhou, J.~Yang, C.~Tan, S.~Pan, and Z.~Zhou, ``Double-image encryption scheme
  combining {DWT-based} compressive sensing with discrete fractional random
  transform,'' \emph{Optics Communications}, 2015.

\bibitem{cui2013image}
D.~Cui, L.~Shu, Y.~Chen, and X.~Wu, ``Image encryption using block based
  transformation with fractional {Fourier} transform,'' in \emph{Proceedings of
  8th International ICST Conference on Communications and Networking in China
  (CHINACOM)}.\hskip 1em plus 0.5em minus 0.4em\relax IEEE, 2013, pp. 552--556.

\bibitem{lang2010image}
J.~Lang, R.~Tao, and Y.~Wang, ``Image encryption based on the
  multiple-parameter discrete fractional {Fourier} transform and chaos
  function,'' \emph{Optics Communications}, vol. 283, no.~10, pp. 2092--2096,
  2010.

\bibitem{qin2009vulnerability}
W.~Qin and X.~Peng, ``Vulnerability to known-plaintext attack of optical
  encryption schemes based on two fractional {Fourier} transform order keys and
  double random phase keys,'' \emph{Journal of Optics A: Pure and Applied
  Optics}, vol.~11, no.~7, p. 075402, 2009.

\bibitem{azoug2016non}
S.~E. Azoug and S.~Bouguezel, ``A non-linear preprocessing for opto-digital
  image encryption using multiple-parameter discrete fractional {Fourier}
  transform,'' \emph{Optics Communications}, vol. 359, pp. 85--94, 2016.

\bibitem{ahmad2015enhanced}
M.~Ahmad, U.~Shamsi, and I.~R. Khan, ``An enhanced image encryption algorithm
  using fractional chaotic systems,'' \emph{Procedia Computer Science},
  vol.~57, pp. 852--859, 2015.

\bibitem{li2015asymmetric}
Y.~Li, F.~Zhang, Y.~Li, and R.~Tao, ``Asymmetric multiple-image encryption
  based on the cascaded fractional {Fourier} transform,'' \emph{Optics and
  Lasers in Engineering}, vol.~72, pp. 18--25, 2015.

\bibitem{li2015modified}
X.~Li and I.~Lee, ``Modified computational integral imaging-based double image
  encryption using fractional {Fourier} transform,'' \emph{Optics and Lasers in
  Engineering}, vol.~66, pp. 112--121, 2015.

\bibitem{singh2009image}
K.~Singh, N.~Singh, P.~Kaur, and R.~Saxena, ``Image compression by using
  fractional transforms,'' in \emph{Proceedings of International Conference on
  Advances in Recent Technologies in Communication and Computing}.\hskip 1em
  plus 0.5em minus 0.4em\relax IEEE, 2009, pp. 411--413.

\bibitem{kumar2013digital}
M.~Kumar, R.~Rewani \emph{et~al.}, ``Digital image watermarking using
  fractional {Fourier} transform via image compression,'' in \emph{Proceedings
  of International Conference on Computational Intelligence and Computing
  Research {(ICCIC)}}.\hskip 1em plus 0.5em minus 0.4em\relax IEEE, 2013, pp.
  1--4.

\bibitem{yetik2001image}
{\.I}.~{\c{S}}. Yetik, M.~A. Kutay, and H.~M. Ozaktas, ``Image representation
  and compression with the fractional {Fourier} transform,'' \emph{Optics
  communications}, vol. 197, no.~4, pp. 275--278, 2001.

\bibitem{kumari2013secure}
P.~V. Kumari and K.~Thanushkodi, ``A secure fast {2D}-discrete fractional
  {Fourier} transform based medical image compression using hybrid encoding
  technique,'' in \emph{Proceedings of International Conference on Current
  Trends in Engineering and Technology {(ICCTET)}}.\hskip 1em plus 0.5em minus
  0.4em\relax IEEE, 2013, pp. 1--7.

\bibitem{zhou2015image}
N.~Zhou, H.~Li, D.~Wang, S.~Pan, and Z.~Zhou, ``Image compression and
  encryption scheme based on {2D} compressive sensing and fractional mellin
  transform,'' \emph{Optics Communications}, vol. 343, pp. 10--21, 2015.

\bibitem{serbes2010optimum}
A.~Serbes and L.~Durak, ``Optimum signal and image recovery by the method of
  alternating projections in fractional {Fourier} domains,''
  \emph{Communications in Nonlinear Science and Numerical Simulation}, vol.~15,
  no.~3, pp. 675--689, 2010.

\bibitem{yang2007blurred}
W.~Yang, Z.~Feng, W.~Liu, and X.~Zou, ``Blurred defocused image restoration
  based on frft,'' \emph{Wuhan University Journal of Natural Sciences},
  vol.~12, no.~3, pp. 496--500, 2007.

\bibitem{paek2013retina}
K.~Paek, M.~Yao, and X.~Xue, ``A retina-filtering model and its application in
  image restoration,'' in \emph{Proceedings of Ninth International Conference
  on Natural Computation {(ICNC)}}.\hskip 1em plus 0.5em minus 0.4em\relax
  IEEE, 2013, pp. 322--327.

\bibitem{chen2014automatic}
K.~Chen, E.~L. Piccolomini, and F.~Zama, ``An automatic regularization
  parameter selection algorithm in the total variation model for image
  deblurring,'' \emph{Numerical Algorithms}, vol.~67, no.~1, pp. 73--92, 2014.

\bibitem{chen2016fractional}
G.~Chen, J.~Zhang, and D.~Li, ``Fractional-order total variation combined with
  sparsifying transforms for compressive sensing sparse image reconstruction,''
  \emph{Journal of Visual Communication and Image Representation}, vol.~38, pp.
  407--422, 2016.

\bibitem{kumar2016fractional}
P.~Kumar, S.~Kumar, and B.~Raman, ``A fractional order variational model for
  the robust estimation of optical flow from image sequences,''
  \emph{Optik-International Journal for Light and Electron Optics}, 2016.

\bibitem{erden1999repeated}
M.~F. Erden, M.~A. Kutay, and H.~M. Ozaktas, ``Repeated filtering in
  consecutive fractional {Fourier} domains and its application to signal
  restoration,'' \emph{IEEE Transactions on Signal Processing}, vol.~47, no.~5,
  pp. 1458--1462, 1999.

\bibitem{yan2001image}
P.~Yan, Y.~L. Mo, and H.~Liu, ``Image restoration based on the discrete
  fraction {Fourier} transform,'' in \emph{Proceedings of Conference on
  Multispectral Image Processing and Pattern Recognition}.\hskip 1em plus 0.5em
  minus 0.4em\relax International Society for Optics and Photonics, 2001, pp.
  280--285.

\bibitem{kutay1998optimal}
M.~A. Kutay and H.~M. Ozaktas, ``Optimal image restoration with the fractional
  {Fourier} transform,'' \emph{JOSA A}, vol.~15, no.~4, pp. 825--833, 1998.

\bibitem{yin2015fractional}
X.~Yin, S.~Zhou, and M.~A. Siddique, ``Fractional nonlinear anisotropic
  diffusion with p-laplace variation method for image restoration,''
  \emph{Multimedia Tools and Applications}, pp. 1--22, 2015.

\end{thebibliography}

 \bigskip \smallskip

 \it

 \noindent
$^1$ School of Mechanical Engineering, \\
ShenYang LiGong University, \\
Shenyang 110159, China \\[4pt]
e-mail: 9501133@163.com    \\[12pt]
$^2$ College of Information Science and Technology, \\
Northeastern University, \\
Shenyang 110819, China\\[4pt]
e-mail: chendali@ise.neu.edu.cn  \\[12pt]
$^3$ School of Engineering, University of California, \\
Merced. 5200 N. Lake Road, \\
Merced, CA 95343, USA \\[4pt]
e-mail: tzhao3@ucmerced.edu   \\[12pt]
$^4$ School of Engineering, University of California, \\
Merced. 5200 N. Lake Road, \\
Merced, CA 95343, USA \\[4pt]
e-mail: ychen53@ucmerced.edu
\hfill Received: MARCH 1, 2016 \\[12pt]
\end{document}